\documentclass{article}

\usepackage[main, final]{neurips_2025}

\usepackage[utf8]{inputenc} 
\usepackage[T1]{fontenc}    
\usepackage{hyperref}       
\usepackage{url}            
\usepackage{booktabs}       
\usepackage{amsfonts}       
\usepackage{amsmath}        
\usepackage{nicefrac}       
\usepackage{microtype}      
\usepackage{xcolor}         
\usepackage{graphicx}       
\usepackage{makecell}       
\usepackage{float}          

\usepackage{enumitem}
\usepackage{booktabs}
\usepackage{multirow}
\usepackage{colortbl}
\usepackage{tabularx}
\usepackage{array}
\usepackage{multirow}
\usepackage[raster,skins, most]{tcolorbox} %

\usepackage[table]{xcolor}

\newcolumntype{Y}{>{\raggedleft\arraybackslash}p{0.12\linewidth}}

\colorlet{red1}{red!20}

\title{REDSearcher: A Scalable and Cost-Efficient Framework for Long-Horizon Search Agents}

\author{%
  REDSearcher Team \\
  \textbf{Project Page:} \href{https://redsearchagent.github.io}{redsearchagent.github.io}
}

\begin{document}

\maketitle

\begin{abstract}
Large language models are transitioning from general-purpose knowledge engines to real-world problem solvers, yet optimizing them for deep search tasks remains challenging. 
The central bottleneck lies in the extreme sparsity of high-quality search trajectories and reward signals, arising from the difficulty of scalable long-horizon task construction and the high cost of interaction-heavy rollouts involving external tool calls.
To address these challenges, we propose REDSearcher, a unified framework that co-designs complex task synthesis, mid-training, and post-training for scalable search-agent optimization.
Specifically, REDSearcher introduces the following improvements: (1) We frame task synthesis as a dual-constrained optimization, where task difficulty is precisely governed by graph topology and evidence dispersion, allowing scalable generation of complex, high-quality tasks. (2) We introduce tool-augmented queries to encourage proactive tool use rather than passive recall.(3) During mid-training, we strengthen core atomic capabilities—knowledge, planning, and function calling—substantially reducing the cost of collecting high-quality trajectories for downstream training. (4) We build a local simulated environment that enables rapid, low-cost algorithmic iteration for reinforcement learning experiments.
Across both text-only and multimodal search-agent benchmarks, our approach achieves state-of-the-art performance. 
To facilitate future research on long-horizon search agents, we will release 10K high-quality complex text search trajectories, 5K multimodal trajectories and 1K text RL query set, and together with code and model checkpoints.
\end{abstract}

\vspace{-0.2cm}
\newcommand{\ours}{REDSearcher}
\begin{figure}[H]
  \centering
  \includegraphics[width=\linewidth]{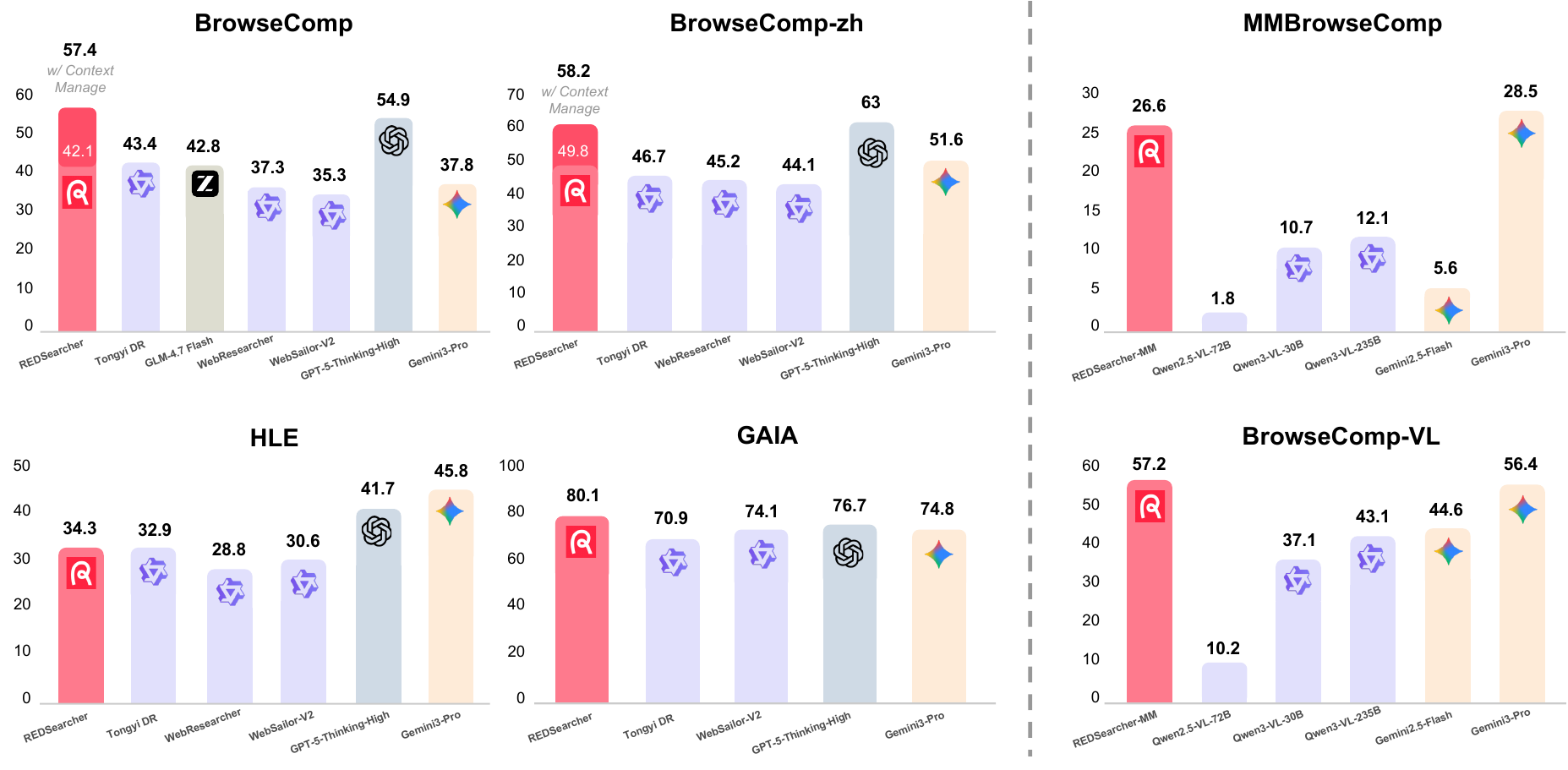}
  \caption{Benchmark performance of REDSearcher.}
  \label{fig:abstract_overview}
\end{figure}

\clearpage

\section{Introduction}
Large language models (LLMs) \cite{achiam2023gpt,touvron2023llama,team2023gemini} are transitioning from static, parametric knowledge engines into dynamic agents \cite{liu2025deepseek,zeng2025glm,team2025kimi} capable of navigating the open world.
While current models excel at simple retrieval tasks \cite{jin2025search}, they struggle with deep search—an interactive, long-horizon setting in which an agent must iteratively acquire evidence, maintain competing hypotheses, and synthesize information across multiple sources. In contrast to standard RAG \cite{arslan2024survey}, which typically relies on static one-shot retrieval, deep search requires closed-loop search-and-reason behavior that adapts to newly found evidence \cite{sun2026deep}.
However, optimizing LLMs for such depth is hindered by a critical bottleneck: the extreme sparsity of effective supervision signals \cite{li2025websailor,li2025websailor2,team2025tongyi}. Scaling these agents is currently intractable due to two prohibitive barriers: the difficulty of synthesizing complex, high-quality reasoning tasks at scale, and the immense computational and temporal cost of collecting interaction-heavy trajectories involving extensive external tool usage.

Accordingly, we propose \textbf{REDSearcher}, a  framework for training tool-augmented deep-search agents across text-only and multimodal (image-text) settings, jointly optimizing task synthesis, mid-training, and post-training to enable \textbf{scalable, controllable, and cost-effective} optimization of long-horizon search behavior.

\textbf{REDSearcher} introduces the following technical components:
\begin{itemize}[leftmargin=0pt]
  \item \textbf{Dual-Constrained Task Synthesis}. We mitigate the scarcity of challenging supervision by formulating query generation as a constraint satisfaction problem over a latent knowledge graph. Unlike standard QA datasets that predominantly admit linear, tree-like reasoning, we construct instances with higher structural complexity (e.g., cycles and interlocking constraints), which increases the effective reasoning load and requires maintaining multiple competing hypotheses rather than simple sequential deduction. In addition, we introduce an explicit \emph{evidence-dispersion} constraint to discourage single-page shortcut solutions: logically coupled facts are deliberately placed in disjoint sources, encouraging iterative planning and cross-document synthesis under realistic search settings.
  \item \textbf{Proactive Tool-Augmented Queries}. Learning to use tools purely via sparse trial-and-error exploration is sample-inefficient. We therefore \emph{tool-ground} the synthesized queries by rewriting key facts into \emph{tool-resolvable constraints} that cannot be satisfied by text retrieval alone. Concretely, we replace explicit entities with operationalized specifications—e.g., turning a place name into a routing/distance constraint resolved by a map tool, or swapping a named entity for a visual cue that requires image understanding. This design makes successful task completion contingent on invoking the appropriate tool, thereby densifying learning signals for targeted tool usage during long-horizon rollouts.
  \item \textbf{Cost-Efficient Mid-Training}. Bridging the gap between static pre-training and dynamic agent deployment requires a dedicated transitional phase. We adopt a two-stage mid-training regimen that separates the acquisition of \emph{atomic subskills} from \emph{interactive execution}. In the first stage, synthetic data strengthens core competencies—intent-anchored grounding (filtering noise to find evidence) and hierarchical planning (structuring ambiguous goals)—at scale without costly environment interaction. The second stage introduces simulated tool-use loops and long-horizon trajectories to capture environmental feedback and state retention. By warm-starting the model with these capabilities before real-world exposure, we significantly improve initial exploration success and reduce the sample complexity and computational cost of collecting high-quality trajectories for downstream training.
  \item \textbf{Functionally Equivalent Simulation Environment}.  To facilitate rapid algorithmic iteration, we construct a lightweight, local simulated environment that mimics real-world web dynamics while eliminating the latency and expense of live API calls. Crucially, this environment is engineered to balance guaranteed solvability with high-interference noise: it ensures that all necessary evidence is present within the closed corpus, yet physically dispersed and buried amidst extensive distractor documents. This design rigorously stress-tests the agent's ability to discriminate valid signals from noise, providing a high-throughput sandbox that enables efficient reinforcement learning experiments and scalable evaluation without the bottlenecks of external network interactions.
\end{itemize}

\section{Preliminary}
\label{sec:02_preliminary}

\subsection{Problem Formulation}
\label{sec:formulation}

We model a web-enabled question answering session as an interactive process between an agent and an environment equipped with external tools.
Let $q$ denote the user question, which may be unimodal (text) or multimodal (e.g., text with an image).
Over multiple steps, the agent issues tool calls, observes returned evidence, and finally produces an answer grounded in the collected information.

\paragraph{Core variables.}
We define the following variables for a session:
\begin{itemize}
  \item \textbf{Question ($q$).} The user-provided information need. In our setting, $q$ can be long, fuzzy, and underspecified, often requiring aggregation across sources and iterative refinement of constraints.
  \item \textbf{Action ($a_t$).} A tool-mediated operation at step $t$ (e.g., issuing a search query, opening a page, following links, extracting snippets, parsing content, deduplicating results, or terminating).
  \item \textbf{Observation ($o_t$).} The tool feedback after executing $a_t$ (e.g., ranked results, snippets, page content, images and associated metadata, and any structured fields produced by tools).
  \item \textbf{Internal state ($\tau_t$).} The agent's working state at step $t$, which serves as a \emph{compact representation} of the interaction history and current constraints (e.g., a reasoning summary, extracted entities/attributes, active hypotheses, and intermediate conclusions) used to decide the next action.
  \item \textbf{Answer ($y$).} The final response produced at the end of the interaction, which should be grounded in collected evidence and satisfy the constraints implied by $q$. When evidence is incomplete or conflicting, $y$ should explicitly reflect uncertainty.
\end{itemize}

\paragraph{Interaction dynamics (fully observed).}
Let $h_t=\big(q,(a_0,o_0),\ldots,(a_{t-1},o_{t-1})\big)$ denote the transcript up to step $t$.
The agent selects the next tool call conditioned on the available context, $a_t\sim\pi(\cdot\mid h_t)$, and receives feedback $o_t=\mathcal{E}(a_t)$.
We treat $\mathcal{E}$ as a deterministic tool interface given the issued request, and any apparent stochasticity (e.g., ranking variability) is absorbed into the returned observation $o_t$.
For multimodal settings, $o_t$ may include images and associated metadata in addition to text, and the transcript $h_t$ aggregates evidence across modalities.
After $T$ steps, the agent outputs $y=g(q,h_T)$.

\subsection{ReAct-style Trajectory Representation}
\label{sec:react}

ReAct~\cite{yao2022react} organizes the interaction as an interleaved sequence of \emph{(state/thought, action, observation)} tuples.
For a single instance, we record the trajectory as
\begin{equation}
\mathcal{H}_T = \big(q,\; (\tau_0, a_0, o_0),\; (\tau_1, a_1, o_1),\; \ldots,\; (\tau_T, a_T, o_T),\; y\big).
\end{equation}
Here, $\tau_t$ summarizes the current constraints and intermediate beliefs derived from the history, $a_t$ is the tool call selected under that state, and $o_t$ is the returned evidence. The final answer $y$ is produced after the last update using the accumulated state and evidence.

\subsection{Context Management}
\label{sec:context_mgmt}

Even with long context windows, search-based agent trajectories can easily grow beyond the model's maximum input length due to repeated tool calls, long webpages, and accumulated intermediate notes.
When the context approaches the window limit, the agent may be forced to truncate earlier steps, which can break constraint tracking and degrade long-horizon performance.
To handle this practical bottleneck, we adopt a simple \emph{context management} strategy: \textbf{Discard-all} \cite{anthropic2025claudeopus45systemcard,liu2025deepseek}.
Concretely, once the running context exceeds a preset threshold of the window budget, we reset the in-context tool-call history (i.e., remove all past $(\tau_i,a_i,o_i)$ pairs from the prompt) while keeping the original question $q$ and a minimal task specification.
The agent then re-initiates the rollout from a fresh context, effectively trading long-term in-context memory for a larger remaining token budget to continue exploration and tool use.

\section{Scalable Complex Task Synthesis}
\label{sec:task_synthesis}

To train deep search agents capable of navigating the open world, we require queries that exhibit specific challenging characteristics: multi-hop reasoning, ambiguity, and non-linear search paths. Solving such queries mandates iterative tool usage and the synthesis of fragmented evidence. However, existing open-source datasets \cite{yang2018hotpotqa,kwiatkowski2019natural} are predominantly constituted of linear, retrieval-friendly tasks that fail to drive the evolution of agentic capabilities. To address this, we establish a scalable, controllable synthesis pipeline.

\subsection{Motivation}
\label{Complexity_motivation}

Before detailing the synthesis pipeline, we first formalize the following question:

\begin{quote}
\textbf{How should the complexity of a deep search problem be characterized?}
\end{quote}

We argue that deep search complexity can be decomposed into two dimensions:
(i) Topological Logical Complexity and 
(ii) Information Source Dispersion.

\subsubsection{Topological Logical Complexity: A Treewidth Perspective}
Reasoning over complex queries can be formulated as constraint satisfaction or traversal problems on an underlying knowledge graph structure. 
A classical insight in algorithmic and database theory is that the computational difficulty of many such graph-structured problems depends critically on \emph{structural} properties of the underlying graph \cite{dalmau2002constraint}.
In particular, while general CSP-style reasoning can be NP-hard, broad families become tractable on instances whose graphs have bounded \textit{treewidth}~\cite{kloks1994treewidth}, and more generally, properties definable in monadic second-order logic admit linear-time algorithms on bounded-treewidth graphs (Courcelle's Theorem)~\cite{courcelle1990monadic}.
This suggests that query difficulty is driven not only by size, but also by how tightly constraints are coupled---e.g., through cycles and limited decomposability.

Motivated by this perspective, we adopt \textbf{treewidth} as a structural metric for topological logical complexity. Let a query's logical structure be represented by a graph $G = (V, E)$. A \textit{tree decomposition} of $G$ is a pair $(T, \{X_i\}_{i \in I})$, where $T$ is a tree and each node $i$ in $T$ is associated with a "bag" of vertices $X_i \subseteq V$, satisfying:
\begin{enumerate}
    \item The union of all bags equals $V$.
    \item For every edge $(u, v) \in E$, there exists a bag $X_i$ containing both $u$ and $v$.
    \item For any vertex $v$, the set of nodes $\{i \mid v \in X_i\}$ forms a connected subtree in $T$.
\end{enumerate}

The width of the decomposition is $\max_{i \in I} |X_i| - 1$. The \textbf{treewidth} of $G$, denoted as $tw(G)$, is the minimum width over all possible tree decompositions of $G$:
\begin{equation}
    tw(G) = \min_{(T, \{X_i\})} \left( \max_{i \in I} |X_i| - 1 \right)
\end{equation}

\paragraph{Complexity Scaling.}
Intuitively, treewidth serves as a proxy for the \emph{working memory} required to satisfy coupled constraints: tree-like structures (low treewidth) admit divide-and-conquer, whereas high treewidth indicates stronger entanglement among variables. As a coarse proxy---consistent with dynamic programming over tree decompositions---we approximate reasoning cost as
\begin{equation}
    \mathcal{C}_{reasoning} \approx O(N \cdot d^{k+1})
\end{equation}
where $N$ is the number of reasoning steps (hops), $d$ is the branching factor per step (e.g., top-$d$ candidates), and $k=tw(G)$.
This highlights that increasing $k$ can impose an exponential burden, forcing the agent to maintain multiple entangled hypotheses rather than performing simple sequential deduction.

\begin{figure}[t]
  \centering
  \includegraphics[width=\linewidth]{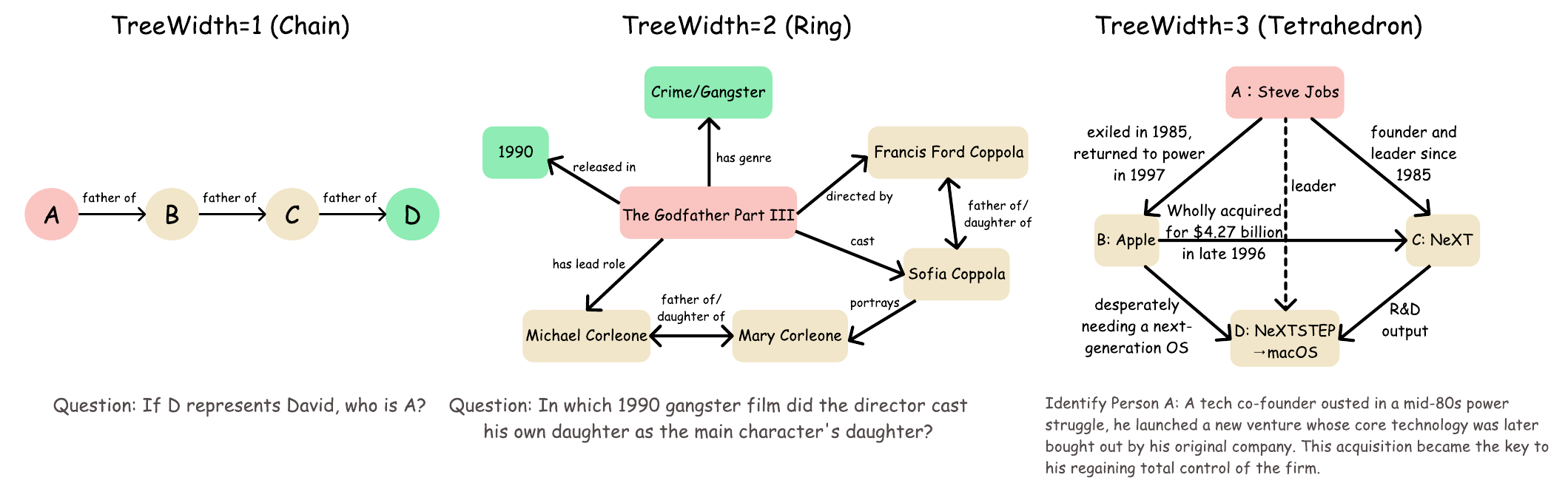}
  \vspace{-5mm}
  \caption{Increasing reasoning complexity as a function of graph treewidth. From left to right, the dependency structure evolves from a simple chain ($k=1$), to a cyclic constraint graph ($k=2$), and finally to a fully coupled tetrahedral structure ($k=3$).
Green nodes denote given entities and red nodes denote the final answer, while yellow nodes represent intermediate reasoning variables. Higher treewidth corresponds to larger jointly maintained variable sets and stronger global consistency constraints, transforming reasoning from linear propagation to high-dimensional constraint satisfaction.}
  \label{fig:tree_width}
\end{figure}

As illustrated in Figure~\ref{fig:tree_width}, we characterize task difficulty through the treewidth $k$ of the underlying reasoning graph. The figure visualizes three representative structural regimes. In each example, green nodes denote observed facts (given entities), yellow nodes correspond to intermediate latent variables that must be inferred, and the red node represents the final answer. As $k$ increases, the structural coupling among variables strengthens, and the reasoning process transitions from simple propagation to globally constrained joint verification.

\begin{itemize}[leftmargin=*]
    \item \textbf{Type I: Linear Reasoning ($k=1$)}. 
    \textit{Structure:} Trees or simple chains.
    \textit{Example:} "A is the father of B, B is the father of C... Who is A?"
    \textit{Cognitive Load:} The agent only needs to track the immediate predecessor. Complexity is polynomial ($O(N \cdot d^2)$). This represents the majority of current multi-hop QA datasets.
    
    \item \textbf{Type II: Cyclic/Diamond Constraints ($k=2$)}. 
    \textit{Structure:} Graphs containing cycles or parallel paths that re-converge.
    \textit{Example:} "In which 1990 gangster film did the director cast his own daughter as the main character's daughter?"
    \textit{Cognitive Load:} The agent must simultaneously satisfy constraints between the \textit{Movie}, \textit{Director}, and \textit{Actress}. This requires maintaining a larger "bag" of variables (triplets) in memory to verify consistency, creating a search space of $O(N \cdot d^3)$. A failure in one branch (e.g., an incorrect daughter) necessitates backtracking.
    
    \item \textbf{Type III: High-Dimensional Coupling ($k \ge 3$)}. 
    \textit{Structure:} Clique-like structures (e.g., Tetrahedron).
    \textit{Example:} "Identify Person A: A tech co-founder ousted in a mid-80s power struggle, he launched a new venture whose core technology was later bought out by his original company. Cognitive Load: Here, variables A, B, C, and D are fully coupled."
    \textit{Cognitive Load:} Here, variables A, B, C, and D are fully coupled. The problem cannot be decomposed into independent sub-problems. The agent must validate a complete $K_4$ subgraph, leading to a combinatorial explosion ($O(N \cdot d^4)$) if effective pruning is not applied.
\end{itemize}

\subsubsection{Distributional Complexity: Information Dispersion}
While treewidth captures the \emph{structural} coupling of a reasoning graph, it does not fully determine search difficulty in open-web settings.
In particular, high information density on the web can create \emph{shortcut retrieval}: a single comprehensive document may contain multiple logically connected facts (e.g., nodes $A,B,C,D$), allowing a theoretically complex instance to be solved with near one-shot retrieval (effectively reducing the required reasoning depth).

To characterize this orthogonal factor, we introduce \textbf{Minimum Source Dispersion (MSD)}, which measures how fragmented the required evidence is across sources.
MSD is defined as the minimum number of distinct documents needed to cover the information required by the reasoning graph $G$:
\begin{equation}
    \mathcal{D}_{task} = \min_{\mathcal{S} \subseteq \mathcal{W}} |\mathcal{S}| \quad \text{s.t.} \quad \text{Cover}(\mathcal{S}, G) = \text{True}
\end{equation}
where $\mathcal{W}$ denotes the document corpus and $\mathcal{S}$ represents a retrieved subset. The condition $\text{Cover}(\mathcal{S}, G)$ implies that the union of information in $\mathcal{S}$ is sufficient to resolve all nodes in graph $G$.

Taken together, \textit{structural} and \textit{distributional} complexity offer a dual view of deep-search difficulty.
In practice, instances are most resistant to shortcut retrieval when both $tw(G)$ and $\mathcal{D}_{\text{task}}$ are high, i.e., when coupled facts are dispersed across disjoint sources.
This motivates our \textbf{dual-constrained optimization} for task synthesis: we jointly control graph topology (treewidth) and evidence dispersion (MSD), encouraging iterative planning and cross-document synthesis under realistic web retrieval.

\subsection{Scalable Complex Task Synthesis Pipeline}
\begin{figure}[ht]
  \centering
  \includegraphics[width=\linewidth]{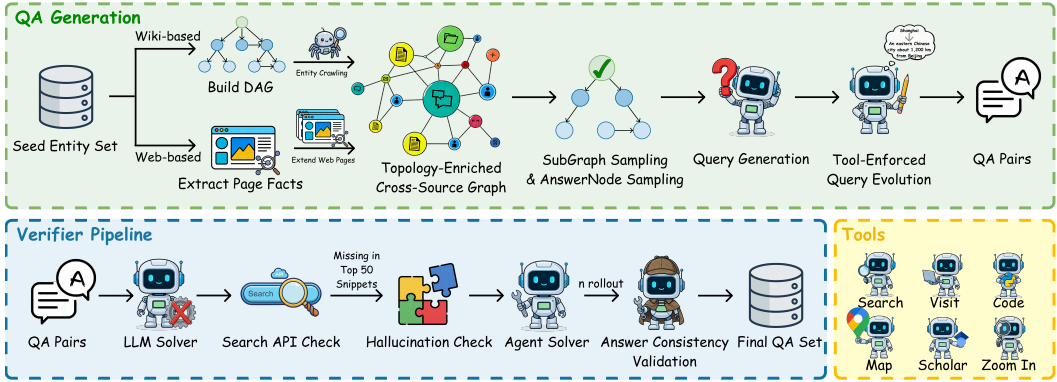}
  \vspace{-4mm}
  \caption{ Overview of the scalable complex task synthesis pipelinee. The process operates via a dual-pathway mechanism to maximize both structural complexity and information dispersion, followed by a rigorous solver-based verification stage.}
  \label{fig:qa_pipeline}
\end{figure}

Guided by the theoretical framework established in \S\ref{Complexity_motivation}, we design a scalable synthesis pipeline to manufacture QA pairs that exhibit specific topological properties (e.g., $k \ge 2$) and high information dispersion. 
As illustrated in Figure~\ref{fig:qa_pipeline}, our pipeline departs from random template filling. Instead, it operates as a \textit{graph-to-text} inverse problem: we first construct a reasoning graph with the desired treewidth and dispersion, and then transform this structure into a natural language query. 
The pipeline consists of two distinct phases: \textit{QA Generation} and \textit{Task Verification}.

\subsubsection{QA generation}
\label{sec:qa_generation_stage}

\paragraph{Seed Collection and Filtering.}
To bootstrap the synthesis pipeline, we initialize a seed pool using English and Chinese Wikipedia entities \cite{vrandevcic2014wikidata}. We apply a filtering cascade to isolate entity-centric pages, pruning noise four criteria: (i) \textbf{main-text length thresholds} to remove pages that are too short (too sparse) or too long (too popular / over-covered); (ii) \textbf{structure filtering} to discard lists, indexes, and glossaries; (iii) \textbf{meta-page removal} for administrative content; and (iv) \textbf{concept filtering}, where an LLM classifier distinguishes concrete entities from abstract theories. We deduplicate aliases and redirects to establish a compact, high-signal seed pool for downstream generation.

\paragraph{Graph Construction and Topological Enrichment.}
We adopt a Directed Acyclic Graph as the fundamental data structure, as it naturally models multi-step reasoning while ensuring auditability.
Starting from a filtered seed entity, we expand the graph through two complementary acquisition streams:
(i) structured relation harvesting from Wikidata, and (ii) hyperlink-based document discovery via web traversal.
These streams run in parallel and serve distinct roles in graph construction, without requiring a unified merge into a single substrate.
Crucially, to transcend simple multi-hop retrieval and achieve the high structural complexity ($k \ge 2$) defined in \S\ref{Complexity_motivation}, we introduce a \textbf{Topology-Enriched Cross-Source Graph} construction phase.
Instead of relying solely on explicit database relations, we deploy an LLM-driven \textit{Graph Agent} to densify the topology.
This densification introduces cycles into the dependency graph, breaking the linearity of search paths. It compels the solver to shift from sequential retrieval to joint constraint satisfaction, where a valid answer is not found by following a single thread, but by verifying that multiple, distributed pieces of evidence are mutually consistent.

\paragraph{Efficient Subgraph and Answer Sampling.} Building a fully enriched, topology-dense graph is computationally expensive, requiring substantial LLM reasoning and retrieval calls. To amortize this cost, we adopt a One-Graph-Multi-Task sampling strategy: from each master graph, we extract multiple distinct connected subgraphs as independent reasoning contexts. Within each subgraph, answer nodes are selected strictly by topological role (e.g., deep leaves vs. high-degree hubs). Different structural positions induce different reasoning requirements (e.g., long-chain backtracking vs. multi-constraint verification), thereby increasing task diversity. Reusing the same underlying graph yields an order-of-magnitude more training instances, effectively distributing the graph construction overhead across dozens of high-quality samples.

\paragraph{Question generation.}
Given the sampled knowledge graph and the target answer, we use a large language model to generate a natural-language question that faithfully captures the graph constraints in a concise, natural form.

\paragraph{Tool-Enforced Query Evolution.}
To enforce the proactive tool-augmented behavior outlined in our motivation, we implement a \textbf{Tool-Injection Strategy} beyond simple text obfuscation.
A specialized \textit{Editor Agent} rewrites each query by converting static entities into tool-resolvable functional dependencies, replacing direct facts with computable constraints.
For example, instead of naming a location, the agent uses a Maps API to specify it via a routing constraint (e.g., ``the city about two hours' drive west of [Entity A]'').
Similarly, a person entity can be substituted with an attribute-based identifier that requires external lookup, such as ``the scholar with approximately $N$ citations'' (or within a narrow citation interval) retrieved from an academic profile index.
These rewrites create informational gaps that cannot be reliably closed by text retrieval alone, making tool execution an intrinsic prerequisite of the reasoning trajectory.

\subsubsection{Verifier pipeline}
\label{sec:verifier_pipeline}

The QA synthesis procedure intentionally increases difficulty (e.g., via fuzzing) and combines signals from multiple local sources (KB and cached webpages).
As a result, a non-trivial fraction of generated instances may become \emph{too easy}, \emph{internally inconsistent} (question--graph--answer mismatch), \emph{weakly retrievable} on the open web, or \emph{non-unique} in their solutions.
To produce a dataset that is both challenging and reliably verifiable, we employ the multi-stage verifier pipeline illustrated in Figure~\ref{fig:qa_pipeline}, which starts with cheap filters and gradually escalates to stronger, more expensive checks:

\begin{enumerate}
  \item \textbf{LLM solver pre-filter (no tools).} We run an LLM solver \emph{without tool access}; if it answers correctly, the instance is treated as insufficiently challenging and removed.
  \item \textbf{Retrievability check (Search snippets).} We query the question with the search engine API; if the given answer does \emph{not} appear in the snippets of the top-$50$ results, we filter the instance as weakly supported for open-web retrieval.
  \item \textbf{Hallucination / inconsistency check.} We provide the grounded evidence used during construction (e.g., KB triples and cached passages) together with the final question to an LLM verifier; instances with clear contradictions are removed.
  \item \textbf{Agent rollout verification.} We run one strong tool-using agents for $n$ independent rollouts; an instance is kept if at least one rollout predicts the given answer, and we record the pass rate as a confidence signal.
  \item \textbf{Answer uniqueness check.} Building on the successful rollouts, we further scrutinize the results for solution multiplicity. 
We discard instances where the agent plausibly identifies valid alternative answers or distinct candidate sets that satisfy the query constraints. 
While not a formal guarantee of uniqueness, this heuristic filter significantly mitigates the risk of ambiguous or underspecified tasks by removing cases where the solver naturally diverges.
\end{enumerate}

\paragraph{Quality study.}
We validate the synthesis pipeline along two axes: \emph{solvability} and \emph{difficulty} under realistic budgets.
First, to assess data fidelity, we perform human verification on a subset of 500 instances. University-level annotators check logical consistency and grounding sufficiency, and over $85\%$ of instances pass verification, indicating that the synthesized problems are well-formed and likely solvable.
Second, to quantify difficulty, we evaluate a strong open model, DeepSeek-V3.2~\cite{liu2025deepseek}, under our standard agent setting, obtaining $\sim40\%$ accuracy.
To further contextualize hardness, we additionally measure time-bounded human solvability: with a 30-minute search budget, annotators solve $47\%$ of instances. Together, these results suggest that our data is largely solvable, yet remains challenging for both models and humans within practical interaction budgets.


\subsection{Multimodal Task Synthesis Pipeline}

\subsubsection{Multimodal QA Generation}
\label{sec:mm_qa_generation}

Our synthesis pipeline can be conveniently migrated to multimodal QA generation.
The key is to reuse the same end-to-end skeleton and only modify a small number of steps to incorporate visual evidence.
This design keeps the dependency structure explicit and verifiable, while allowing us to scale multimodal synthesis with nearly the same efficiency as text-only synthesis.
As a result, the multimodal pipeline inherits the same desirable properties as the text-only setting: scalability, controllable difficulty, explicit dependencies, and verifiability.

Concretely, we introduce \emph{modality injection} to turn a purely textual reasoning DAG into a cross-modal reasoning DAG, where some constraints are anchored in images and must be resolved via visual understanding.
We then extend fuzzing and verification with image-aware variants (\S\ref{sec:mm_qa_generation}), so that the resulting multimodal questions remain challenging and grounded.

\paragraph{Modality injection.}

We implement modality injection via two complementary mechanisms.
\textbf{Visual attribute anchoring} selects an intermediate node $u$ in the DAG and augments its attribute field with an \emph{image-grounded textual description}.
Concretely, we attach an image to node $u$ and generate (or retrieve from cached pages) a detailed textual description of the visual content (e.g., salient objects, scene type, distinctive symbols, or chart patterns).
This description is stored as part of the node attributes and is treated as a constraint for downstream construction, enabling the question to reference visual evidence without revealing the final answer.
\textbf{Cross-modal dependency} enforces a \emph{visual irreplaceability} constraint: without extracting the required visual cue from the image (e.g., a background object, an emblem on clothing, or a trend line in a chart), the model cannot obtain the information needed to derive the downstream node $v$.
This prevents the image from being decorative and ensures that successful solving requires both visual understanding and external search.

\paragraph{Multimodal question fuzzing.}
We introduce image-aware fuzzing strategies.
\textbf{Visual-semantic abstraction} avoids directly naming the image content in the question and instead uses abstract references (e.g., pronouns or relative descriptions), forcing the model to first recognize the visual entity and then search.
\textbf{Modality translation} allows visual evidence to be injected at arbitrary positions along the reasoning trajectory, rather than only at the beginning.
By replacing selected intermediate textual constraints with image-grounded descriptions, we can (i) place a ``visual bottleneck'' after several text-based steps to increase effective reasoning depth, and (ii) control difficulty more finely by choosing which intermediate constraint must be resolved visually.

\paragraph{Multimodal verifier pipeline.}
We build the multimodal verifier pipeline by starting from the text-only verifier (\S\ref{sec:verifier_pipeline}) and adding extra checks to ensure that the image is both necessary and consistent.
In particular, we remove instances that remain solvable without using vision: \textbf{text-only solvability} discards cases where a pure-text reasoner can answer correctly, and \textbf{text-only retrievability} discards cases where a text-only web-search agent can recover the answer without accessing the image.

The multimodal setting also introduces additional failure modes (e.g., images being too revealing or irrelevant).
We therefore further extend the verifier pipeline with visual-consistency checks.
\textbf{Vision-only solvability check} runs a vision-language model with image input only; if the answer can be guessed from the image without search, the instance is discarded as overly direct.
\textbf{Visual-search alignment} verifies that the image content and retrieved webpages form a complementary reasoning loop, filtering instances where the image is unrelated or purely decorative.
\textbf{Multimodal agent rollout} evaluates a vision-capable tool-using agent end-to-end and records its success rate; instances with consistently high success rate over multiple rollouts are considered too easy and discarded.

By integrating modality injection with vision-aware verification, our pipeline turns a static knowledge graph into a dynamic cross-modal reasoning scaffold.
This design ensures that the resulting multimodal QA pairs are not merely text questions accompanied by decorative images, but visually grounded search tasks that require tight coupling between perception, reasoning, and retrieval.
Importantly, this multimodal extension only requires simple yet necessary modifications to the original synthesis pipeline, enabling efficient large-scale multimodal QA generation.

\subsubsection{Multimodal Trajectory Generation}

We synthesize high-quality SFT trajectories using a ReAct~\cite{yao2022react} agent instantiated with standardized tool schemas. Qwen3VL-235B~\cite{bai2025qwen3vltechnicalreport} alternates between generating intent-aware reasoning and issuing structured tool calls; tool outputs are returned as observations to guide subsequent steps. For efficiency and stability, we cap each episode at 20 interaction rounds, after which the model must produce a final answer. We retain only trajectories whose final answers match the ground-truth labels for supervised fine-tuning.

\section{Overall Training Recipe}
\label{sec:training_recipe}

\begin{figure}[t]
\vspace{-6mm}
  \centering
  \includegraphics[width=\linewidth]{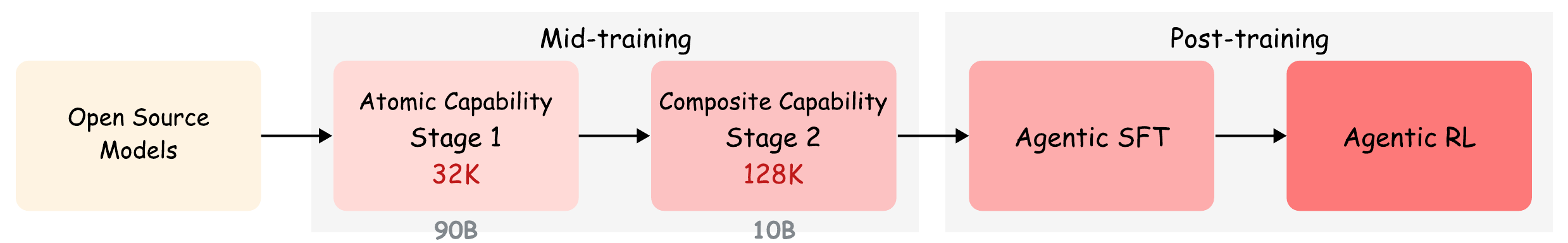}
  \caption{Mid-training and post-training stages for REDSearcher.}
  \label{fig:training_pipeline}
\end{figure}

We start from pretrained open source models and specialize it for multi-turn online web search with tool interaction.
Our training follows a two-phase recipe, as shown in Figure~\ref{fig:training_pipeline}.
\textbf{Mid-training} exposes the model to long-horizon search traces and tool-use patterns, leveraging large-scale synthetic data to ensure sufficient coverage of diverse reasoning trajectories, so that it learns stable interleaved behaviors without degrading its general language ability at low cost.
\textbf{Post-training} subsequently optimizes end-to-end behavior, thereby enhancing the model's agentic reasoning capabilities for complex information seeking.

\section{Agentic Mid-Training via Low-Cost Large-Scale Data Synthesis}

\begin{figure}[ht]
  \centering
  \includegraphics[width=\linewidth]{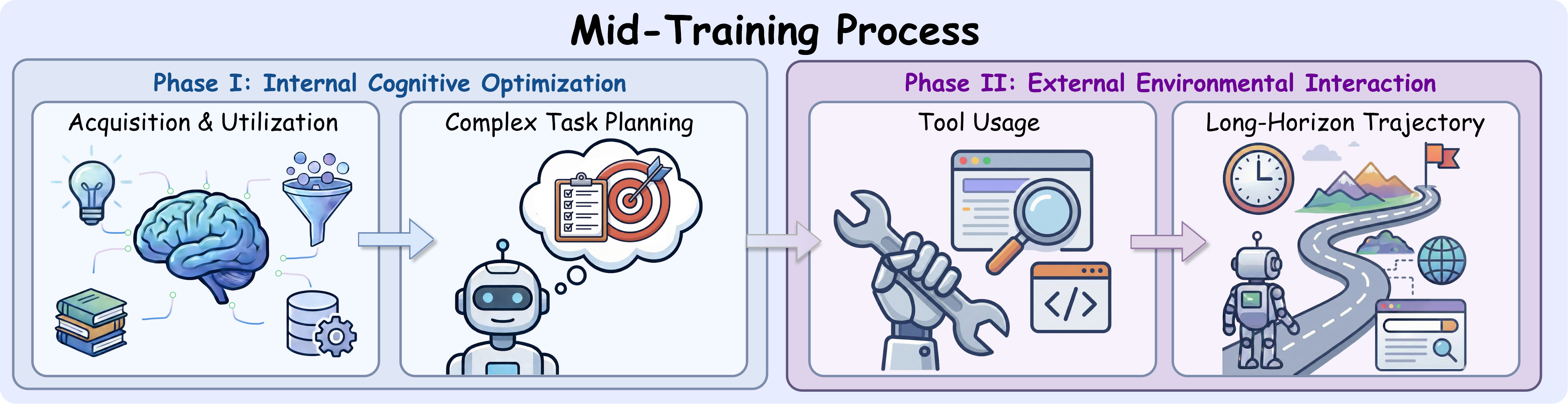}
  \caption{Two stage agentic mid-training framework.}
  \label{fig:mid_train}
\end{figure}

While pre-training equips LLMs with strong knowledge and reasoning capabilities, it lacks experiential interaction with external environments, leaving a pronounced capability gap for agentic tasks requiring environmental perception, action execution, and feedback-driven strategy refinement. 
To bridge this gap, we introduce agentic mid-training as a critical bridge between general-purpose pre-training and agent-specific post-training, comprising two sequential phases: 
the first strengthens atomic capabilities, including knowledge grounding and planning; 
the second builds upon this foundation to develop multi-turn environmental interaction and long-horizon decision-making capabilities.

However, acquiring large-scale mid-training data through manual annotation or real-world environment interaction is prohibitively expensive. To address this, we propose a scalable and cost-effective data synthesis framework for generating agent mid-training data at scale.



\subsection{Stage I: Intent-anchored Grounding and Hierachical Planning (32K Context)}

Search-Agent tasks necessitate that models plan multi-step search strategies throughout long-horizon interactions and filter as well as integrate information from a substantial volume of web pages. This process fundamentally relies upon two core atomic capabilities: the Grounding capability, which facilitates the extraction of key information from redundant observations in accordance with current intent, and the Hierarchical Planning capability, which decomposes complex tasks into hierarchical sub-goals to support multi-step planning while maintaining alignment with global objectives.

\paragraph{Intent-anchored Grounding}

Within deep search tasks, models are required to accurately identify information that is absent from the current reasoning step amidst noisy web browsing environments. We refer to this process as Intent-anchored Grounding. This step serves as the cornerstone of deep search agents; it is imperative that we ensure models acquire accurate and comprehensive information during this stage while avoiding the generation of hallucinations, thus laying a robust foundation for subsequent long-horizon search tasks.

To accomplish this, we adopt a reverse question-answer synthesis approach incorporating distractors.
More specifically, given a central entity $\mathcal{E}$ along with its corresponding document $\mathcal{D}$, we extract factual segments $\mathcal{F}$ pertinent to the central entity from the document, which encompass related events and attributes associated with the central entity. Following this extraction, we synthesize query intents $\mathcal{Q}$ related to the central entity based on these factual segments. Through this methodology, we are able to establish correspondences between the central entity under varying query intents.
Moreover, in order to adapt to the noisy characteristics of web search environments, our input documents incorporate not only documents relevant to the central entity but also irrelevant distractor documents, which collectively serve as the final input. 
We leverage publicly available Wikipedia dumps and cached web crawls as seeds for QA synthesis, requiring no additional data collection effort.

\paragraph{Hierarchical Planning}

When confronted with complex tasks, planning capability assumes paramount importance. In contrast to conventional multi-turn question-answering problems that possess clearly defined reasoning structures, deep search problems tend to exhibit greater ambiguity and necessitate an increased number of reasoning hops. 
It is unrealistic to directly determine each subsequent search and reasoning step based solely on the initial problem formulation.

To address this challenge, we propose to resolve this issue through hierarchical planning.
Hierarchical planning partitions the entry points for solving complex problems into two distinct categories: concrete goals that currently require explicit resolution (for instance, possessing a clear query intent and desiring to obtain specific information related to that query intent), 
and ambiguous goals that require resolution in the future (seeking to narrow uncertainty through queries in order to determine a specific target). 
This partitioning renders long-horizon planning for complex problems feasible, enabling the model to maintain awareness throughout the search process regarding both information that has already been acquired and information that remains to be obtained.
We leverage the topological structure information of knowledge base entities and web pages that is obtained during the QAs synthesis pipeline. 
We flatten the graph along the information flow direction, then leverage LLMs to generate corresponding plans based on the preceding context.

\subsection{Stage II: Agentic Tool Use and Long-horizon Interaction (128K Context)}

Pre-trained models lack exposure to environmental feedback—a critical component in agent systems. To address this, we incorporate tool-calling data involving external environment interactions during Mid-Training. 
We further introduce long-horizon interaction trajectories to strengthen the model's agentic capabilities in complex deep information seeking scenarios.

However, acquiring large-scale observations and trajectories through real-world environment is prohibitively expensive. 
To address this, we adopt two cost-effective and scalable strategies: 
(1) leveraging LLMs to generate a large number of tool protocols and simulate diverse tool-calling interactions without invoking external services, and 
(2) deploying simulated environments to efficiently collect long-horizon agentic interaction trajectories.

\paragraph{Agentic Tool Use}

To enable the model to acquire the capability of perceiving and responding to environmental feedback, we construct multi-turn tool-calling data that encompasses complete ReACT loops.
However, invoking external tools, such as web search operations and external APIs, incurs substantial costs. 
To this end, we employ simulated environments to achieve large-scale environment augmentation during the Mid-Training stage.
We utilize LLMs to generate tool sets, which include tool descriptions, interface signatures, and tool invocation chains. 
Subsequently, we synthesize relevant queries based on these tool sets and employ LLMs to provide environmental feedback for tool invocations. 
This approach enables the collection of extensive multi-turn tool interaction trajectories with diverse tool-calling patterns at scale.

\paragraph{Long-Horizon Interaction}

Deep search tasks often involve dozens of search iterations, during which the model confronts core challenges including state space explosion, historical information forgetting, and goal consistency maintenance. To address these challenges, we introduce long-horizon environmental interaction data to enhance the model's optimization in long-context scenarios.

In long-horizon search scenarios, using LLMs to simulate environmental inputs becomes infeasible, both from cost considerations and from the perspective of generation correctness and consistency.
To overcome this limitation, we construct a comprehensive local simulated web search environment based on Wikipedia and Web Crawl Dumps, supporting fundamental web search and webpage access operations. Moreover, our comprehensive simulated web search environment ensures that the complex queries synthesized through our data pipeline are solvable within the local environment.
We employ large-scale synthesized complex queries as inputs to generate trajectories within the local search environment, which are utilized to enhance the model's long-horizon reasoning capabilities.

\section{Agentic Post-Training}

Through our mid-training phase, the model has acquired foundational capabilities for agentic tasks.
In the post-training phase, we aim to activate these capabilities using high-quality data to enhance the model's performance on downstream deep search tasks. 

\ours{} employs a two-stage post-training process: supervised fine-tuning on synthesized agentic trajectories, followed by agentic reinforcement learning.

\subsection{High-quality Trajectory Synthesis in Real-world Environments}

\paragraph{Real-world Environment Interface}

\ours{} employs five real-world environment interfaces, including web search, web visit, python code execution, google scholar, and google maps.

\textbf{Search} uses google search engine for information retrieval from the Internet. The interface accepts multiple queries as input, and returns a list of organic results, including page title, snippet, and url.

\textbf{Visit} is used to access specific information in a url. The interface accepts a url and a goal as input, and return the web page by Jina. Typically, we use a summarizer to summarize webpage information according to the goal to alleviate the context pressure on the agent model.

\textbf{Python} provides agents with a code sandbox execution environment, supporting tasks such as mathematical calculations, data processing, and logical reasoning. Agents can write and execute Python code and obtain execution results. 

\textbf{Google Scholar} is specifically designed for academic literature retrieval, supporting agents in searching for academic papers, citation information, and author profiles

\textbf{Google Maps} provides geographic location and map-related services, including place search, route planning, distance calculation, and geographic information queries. This interface enables agents to handle tasks involving spatial reasoning and geographic knowledge.

\paragraph{High-quality Trajectory Synthesis}

We develop a low-cost, highly scalable framework for generating complex deep search questions, as illustrated in Figure~\ref{fig:qa_pipeline}.
This framework enables automated large-scale data synthesis at minimal cost without human intervention, while the synthesized problems achieve difficulty levels comparable to BrowseComp.\footnote{DeepSeek-V3.2 achieves a average@4 of approximately 40\% on our synthetic QA dataset.} 
In our trajectory synthesis and agentic reinforcement learning process, we exclusively use QAs synthesized through our own pipeline.

We employ the ReAct workflow for trajectory synthesis. This paradigm addresses complex problems through an iterative \textit{thought-action-observation} loop: at each turn, the agent makes decisions and invokes tools based on prior context, receives observations from the environment, and repeats this process until a final answer is produced. During synthesis, we set the maximum context length to 128K tokens. 
Samples exceeding the maximum length are discarded rather than forcing a response.

Post-filtering is applied to ensure the correctness of trajectories used for supervised fine-tuning.
First, we retain only samples where the final answer is correct.
Second, to prevent the model from learning incorrect patterns or behaviors, we filter out samples that contain a substantial number of failed action and tool response.
Finally, to promote sample diversity, we preserve only one trajectory per question.

\subsection{Supervised Fine-tuning}

We conduct supervised fine-tuning (SFT) on the mid-training checkpoint using high-quality trajectories to enhance RedSearcher's agentic reasoning capabilities. 
During this stage, we employ the standard next-token prediction loss while masking the environment observation portions to exclude them from gradient updates.
We set the maximum context length to 128K during SFT.

\subsection{Agentic Reinforcement Learning}

We employ reinforcement learning with verifiable rewards (RLVR) to enable the continuous improvement of the policy agent through interactions with real environments. 
The policy model interacts with the environment through the ReAct (Reasoning and Acting) paradigm. 
At each turn, the model generates thoughts and executes corresponding actions, then adjusts its subsequent strategy based on environmental observation. 
After rollouts conclude, the LLM judge provides a verifiable reward by evaluating the alignment between the agent's prediction and the ground-truth answer.

\paragraph{RL Algorithm.} We use GRPO~\citep{shao2024deepseekmath} as the training algorithm during reinforcement learning. 
Concretely, for each question we sample a group of trajectories, compute their final rewards, and normalize rewards within the group to obtain relative advantages.
We update $\pi_\theta$ with a clipped policy-gradient objective using these relative advantages.
Following DAPO~\citep{dapo}, we use clip higher during training.
The final reward $\{0/1\}$ only indicates the correctness of the model prediction, and since the model has already learned the required format during SFT, we do not employ any format rewards.
\begin{equation}
  \mathcal{J}_{\mathrm{GRPO}}(\theta)
  =
  \mathbb{E}_{q}\!\left[
    \frac{1}{K}\sum_{k=1}^{K}
    \min\!\Big(
      \rho_{q,k}(\theta)\,\hat{A}_{q,k},\;
      \mathrm{clip}\big(\rho_{q,k}(\theta), 1-\epsilon, 1+\epsilon\big)\,\hat{A}_{q,k}
    \Big)
  \right],
\end{equation}

where $K$ is the number of rollouts per question and $\mathcal{H}^k_T=(q,\tau^k_0,a^k_0,o^k_0,\ldots,\tau^k_T,y^k)$ denotes the $k$-th rollout trajectory under ReACT paradigm.
The advantage $\hat{A}_{q,k}$ of k-th sample of $q$ is computed via group-relative normalization:
\begin{equation}
  \hat{A}_{q,k} = \frac{r_{q,k} - \bar{r}_q}{\sigma_q + \epsilon},
  \quad
  \bar{r}_q = \frac{1}{K}\sum_{k=1}^{K} r_{q,k},
  \quad
  \sigma_q = \sqrt{\frac{1}{K}\sum_{k=1}^{K}(r_{q,k} - \bar{r}_q)^2},
\end{equation}
where $r_{q,k} \in \{0, 1\}$ denotes the outcome reward for the $k$-th trajectory.



\paragraph{Functionally Equivalent Simulation Environment.}

Real-world web search APIs pose several challenges in early-stage experiments, such as unstable external interfaces\footnote{Web crawling tools often suffer from high failure rates due to network instability and access restrictions.} and high query overhead, which hinder rapid experiment iteration.
To address this, we construct an offline simulated search environment.
When constructing the simulated environment, we focus on three key aspects:
\begin{itemize}
    \item \textbf{Interface Consistency}: The API specifications should remain consistent with real search APIs to ensure experimental results are transferable to real-world settings.
    
    \item \textbf{Evidence Completeness}: The simulated environment should encompass all essential evidence required to answer the synthetic queries, including both directly supporting snippets and intermediate evidence necessary for multi-hop reasoning.
    
    \item \textbf{Environmental Noise}: The simulated environment should not be overly simplistic; it should be sufficiently large in scale and incorporate adequate distracting information to simulate the inherent noise and uncertainty in real-world web search scenario.
\end{itemize}

To this end, we construct a large-scale local search environment containing tens of millions of documents. This environment is built upon finewiki dumps and cached web search and visit results collected during the QAs synthesis process.
Our environment supports three commonly used tools in search tasks: search, visit, and python.
To prevent the model from being biased by Wikipedia's URL patterns, we implement a URL obfuscation pipeline. 
Specifically, we construct a URL template library categorized by entity domain, then leverage an LLM to identify the domain of each entity given the snippet and sample a synthetic URL from the corresponding templates.
The search contents retrieved during our data construction pipeline are already cached in the local search repository, thereby ensuring the completeness of the local environment for solving synthesized questions.
Moreover, the tens of millions of documents also ensure a sufficient level of noise in the environment, preventing the model from developing biased capabilities due to an overly simplistic setting.

\paragraph{RL Query Curation}
For the RL query set construction, we filter out samples that are either too simple or too difficult, as these samples fail to provide effective learning signals during training. Our query set is derived from diverse synthesis pipelines, thereby covering a wide range of problem-solving patterns and difficulty gradients.
Furthermore, we observe that automatically constructed QAs often suffers from issues such as multiple valid answers or inconsistent ground truth, which can severely interfere with the learning signals in RLVR with outcome-based rewards.

To address this, we introduce an Agent-as-Verifier pipeline, where a verifier agent retrieves relevant information through external tool calls and compares it against the question's metadata and trajectory to determine the validity of each question.
Human evaluation results demonstrate that this pipeline reduces the error rate of the RL query set to merely 10\% of the original.

\paragraph{RL Training Framework}
During the rollout phase, the agent needs to interact extensively with the environment.
Traditional synchronous rollout approaches significantly slow down training efficiency. 
To address this, we implement an asynchronous rollout workflow based on Slime~\citep{slime_github}, effectively improving rollout throughput.
Furthermore, rollout lengths in deep search tasks often reach up to 128k tokens, making efficient prefix cache hits critical for rollout performance.
To tackle this problem, we design a two-tier rollout load balancing strategy: 
requests within the same rollout maintain inference engine affinity to maximize prefix cache reuse, 
while load balancing across inference engines is achieved through a combination of round-robin and least-access scheduling.

For environment interaction, we deploy a dedicated server to handle external environment calls during RL training. 
This server encapsulates all tool call interfaces into unified request interfaces and implements fallback strategies for error-prone interfaces such as search and web crawling, thereby ensuring maximum stability of environment interactions throughout the training process.

\section{Experiments}
\label{sec:experiments}

\subsection{Experimental Setup}

\paragraph{Benchmarks.}
Following prior work, we evaluate our model on a diverse set of highly challenging benchmarks and compare against representative baselines.
We adhere to each benchmark's official evaluation protocol.
Our evaluation suite includes: Humanity's Last Exam~\cite{phan2025humanity}, BrowseComp~\cite{wei2025browsecomp}, BrowseComp-ZH~\cite{zhou2025browsecomp}, GAIA~\cite{mialon2023gaia}.

We also evaluate on multimodal search benchmarks to validate our strong multimodal retrieval and reasoning capability, including MM-BrowseComp~\cite{li2025mm}, BrowseComp-VL~\cite{geng2025webwatcher}, MMSearch-Plus~\cite{tao2025mmsearch}, MMSearch~\cite{wu2025mmsearch}, and LiveVQA~\cite{fu2025livevqa}.

\paragraph{Baselines.}
We compare our model with the strongest existing search-agent baselines, including (1) proprietary agents, such as Seed1.8~\cite{seedseed1}, Gemini-3-Pro~\cite{gemini3pro2025}, GPT-5.2~\cite{singh2025openai}; (2) open-source agents, such as Kimi-K2.5~\cite{team2026kimi}, GLM-4.7~\cite{zeng2025glm}, DeepSeek-V3.2~\cite{liu2025deepseek}; (3) open-source lightweight agents, including Tongyi DeepResearch~\cite{team2025tongyi}, GLM-4.7-Flash~\cite{zeng2025glm}, and so on.

We also compare against state-of-the-art multimodal search models, including Gemini-3-Pro~\cite{gemini3pro2025}, Seed1.8~\cite{seedseed1}, and an agent workflow built on Qwen3-VL~\cite{bai2025qwen3vltechnicalreport} with the same toolset as used in our experiments. Besides, we also compare \ours{}-MM with existing multimodal deepsearch agents, such as DeepEyesV2~\cite{hong2025deepeyesv2} and Vision-DeepResearch~\cite{huang2026vision}.

\paragraph{Implementation Details.}
Full implementation details for reproducibility are deferred to Appendix~\ref{Sec:Implementation details}.

\subsection{LLM Experimental Results}

\subsubsection{Main Results}
As presented in Table ~\ref{tab:main_results_updated} , \ours{} establishes a new state-of-the-art among open-source agents in the 30B parameter class. With the integration of our context management technique, the model achieves an Overall score of 51.3, substantially outperforming leading same-scale competitors such as Tongyi DeepResearch-30B (48.5)~\cite{team2025tongyi} and WebSailorV2-30B (46.0)~\cite{li2025websailor2}. Beyond its dominance in the open-source landscape, REDSearcher exhibits remarkable competitiveness against larger proprietary models. It surpasses both Claude-4.5-sonnet (41.1)~\cite{claude45} and OpenAI-o3 (49.6)~\cite{o3} in overall performance metrics. Most strikingly on the GAIA benchmark, which evaluates complex agentic capabilities, REDSearcher attains a score of 80.1, outstripping even the GPT-5–Thinking–high model (76.7)~\cite{singh2025openai}. These results underscore the efficacy of our proposed architecture, demonstrating that REDSearcher delivers top-tier deep research capabilities with superior parameter efficiency.

\begin{table}[t]
\centering
\caption{Comparison between REDSearcher and closed / open agentic models. The performance with the context management technique is noted with $^{*}$.}
\label{tab:main_results_updated}
\scriptsize
\setlength{\tabcolsep}{4pt} 
\renewcommand{\arraystretch}{1.15} 

\resizebox{\linewidth}{!}{%
\begin{tabular}{lcccccc}
\toprule
\textbf{Backbone} & \textbf{Size} & \textbf{BrowseComp}~\cite{wei2025browsecomp} & \textbf{BrowseComp-zh}~\cite{zhou2025browsecomp} & \textbf{GAIA}~\cite{mialon2023gaia} & \textbf{HLE}~\cite{phan2025humanity} & \textbf{Overall} \\
\midrule

\rowcolor{gray!20}
\multicolumn{7}{c}{\textbf{\tiny Proprietary Deep Research Agents}} \\
\midrule
Seed1.8~\cite{seedseed1} & - & 67.6 & 81.3 & 87.4 & 40.9 & 69.3\\
Gemini–2.5–pro–DR~\cite{comanici2025gemini} & - & 7.6 & 27.3 & - & - & - \\
Gemini–3–Pro~\cite{gemini3pro2025} & - & 37.8 & 51.6 & 74.8 & 45.8 & 52.5 \\
Claude–4.5–sonnet~\cite{claude45} & - & 24.1 & 42.4 & 66.0 & 32.0 & 41.1 \\
OpenAI–o3~\cite{o3} & - & 49.7 & 58.1 & 70.5 & 20.2 & 49.6 \\
GPT–5–Thinking–high~\cite{singh2025openai} & - & 54.9 & 63.0 & 76.7 & 41.7 & 59.1 \\
GPT–5.2–Thinking–xhigh~\cite{singh2025openai} & - & 65.8 & 76.1 & - & - & - \\

\midrule
\rowcolor{gray!20}
\multicolumn{7}{c}{\textbf{\tiny Open-source Deep Research Agents}} \\
\midrule
Kimi–K2.5–Agent~\cite{team2026kimi} & 1T–A32B & 60.6 / 74.9$^{*}$   & - & - & 50.2 & - \\
GLM–4.7~\cite{zeng2025glm} & 355B–A32B & 52.0 / 66.6$^{*}$  &  - / 67.5$^{*}$ & - & 42.8 \\
DeepSeek–V3.2~\cite{liu2025deepseek} & 671B–A37B & 51.4 / 67.6$^{*}$ & - / 65.0$^{*}$ & - & 40.8 & - \\
LongCat–Flash–Thinking~\cite{team2026longcat} & 560B–A27B & 56.6 / 73.1$^{*}$  & 69.0 / 77.7$^{*}$ & - & - & - \\

\midrule
\rowcolor{gray!20}
\multicolumn{7}{c}{\textbf{\tiny Open-source 30B–A3B Agents}} \\
\midrule
WebResearcher–30B~\cite{qiao2025webresearcher} & 30B–A3B & 37.3 & 45.2 & - & 28.8 & - \\
WebSailorV2–30B~\cite{li2025websailor} & 30B–A3B & 35.3 & 44.1 & 74.1 & 30.6 & 46.0 \\
Tongyi DeepResearch–30B~\cite{team2025tongyi} & 30B–A3B & 43.4 & 46.7 & 70.9 & 32.9 & 48.5 \\
GLM–4.7–Flash~\cite{zeng2025glm} & 30B–A3B & 42.8 & - & - & - & - \\
\midrule
\rowcolor{pink!30}
REDSearcher & 30B–A3B & 42.1 / \textbf{57.4}$^{*}$ & 49.8 / \textbf{58.2}$^{*}$ & \textbf{80.1} & \textbf{34.3} & \textbf{51.6}\\

\bottomrule
\end{tabular}
}

\end{table}

\subsubsection{Ablation of Mid-Training Stages}

Table~\ref{tab:performance} summarizes the progressive impact of the mid-training stages. Overall, we observe a steady improvement in average performance (42.81 to 47.39), validating mid-training as a critical bridge for developing agentic capabilities.

\textbf{Stage I (Grounding \& Planning)} focuses on building atomic competencies. The introduction of Intent-anchored Grounding improves BrowseComp (+1.87) by enhancing information extraction from noisy environments. Furthermore, Hierarchical Planning leads to a significant leap in GAIA (+4.13), confirming that partitioning goals into concrete and ambiguous sub-tasks is essential for complex reasoning.

\textbf{Stage II (Agentic Tool Use \& Interaction)} facilitates the transition from "understanding" to "acting." By incorporating environmental feedback and long-horizon trajectories, we see the most substantial gains in BrowseComp-ZH (+8.91). This breakthrough demonstrates that exposure to real-world action-feedback loops and 128K context is crucial for maintaining goal consistency and robust execution in deep search scenarios.

\begin{table}[htbp]
\centering
\vspace{-5mm}
\caption{Effect of progressive mid-training stages on downstream SFT performance across four benchmarks. Each stage builds upon the previous one to incrementally improve model capabilities.}
\label{tab:performance}
\begin{tabular}{lcccc}
\toprule
 & \textbf{Base} & \textbf{Stage I. Grounding} & \textbf{Stage I. Planning} & \textbf{Stage II. Agentic} \\
\midrule
BrowseComp & 34.74 & 36.61 & 36.97 & 40.44 \\
BrowseComp-ZH & 26.82 & 27.34 & 29.84 & 38.75 \\
Human Last Exam & 32.25 & 32.00 & 31.37 & 31.25 \\
GAIA & 77.43 & 76.70 & 80.83 & 79.13 \\
\midrule
\textbf{Average} & 42.81 & 43.16 & 44.75 & 47.39 \\
\bottomrule
\end{tabular}
\end{table}

\subsection{RL Continues to Advance Model Capabilities}
\begin{figure}[t]
  \centering
  \includegraphics[width=1.0\linewidth]{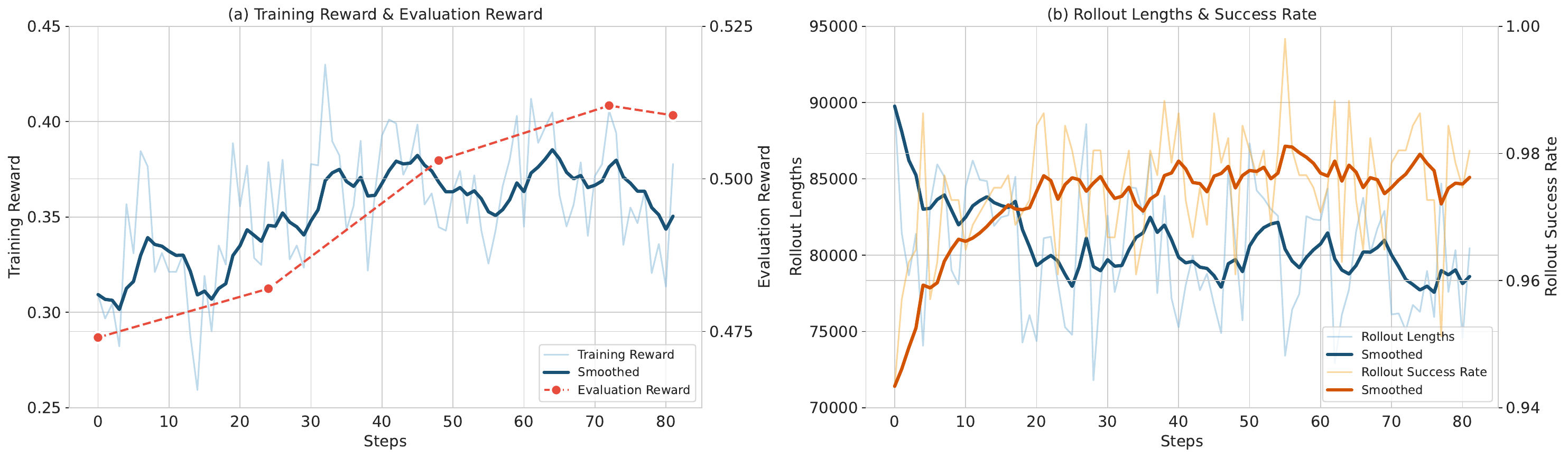}
  \caption{Training dynamics of \ours{} during Agentic Reinforcement Learning. 
    \textbf{(a)} Training reward and evaluation reward across training steps. Evaluation reward is computed over BC, BC-ZH, HLE, and GAIA benchmarks.
    \textbf{(b)} Rollout lengths and rollout success rate during training. }
  \label{fig:rl_dynamics}
\end{figure}

We investigate the effectiveness of agentic RL in enhancing the long-horizon search capabilities of LLMs. 
As shown in Figure~\ref{fig:rl_dynamics}, the model's performance continuously improves with RL training.

As shown in Figure~\ref{fig:rl_dynamics} (a), agentic RL continues to yield consistent improvements even when initialized from a relatively strong SFT checkpoint. 
Prior to RL training, the SFT model achieves an average evaluation reward of 47.4 across four benchmarks (BrowseComp, BrowseComp-zh, HLE, and GAIA), with a BrowseComp score of 39.4.
Following RL training, the average reward increases to 51.3 (\textcolor{green}{+3.9}) and the BrowseComp score rises to 42.1 (\textcolor{green}{+2.7}), corresponding to a relative performance gain of approximately 8.2\% and 6.8\%, respectively.

Besides, we observe an interesting trend in search efficiency during training.
As shown in Figure~\ref{fig:rl_dynamics} (b), the rollout length gradually decreases over RL training, while the reward remains stable or continues to improve. 
This phenomenon suggests that the model learns more efficient explore and search strategies through RL. 
Quantitatively, the average number of tool calls decreases from 100.6 to 90.1, representing a 10.4\% reduction. 
The fact that performance does not degrade despite shorter trajectories indicates that the model has learned to identify more streamlined strategies for task completion, minimizing redundant tool calls without sacrificing effectiveness.

\subsubsection{Decoupling Tool Use from Parametric Knowledge}
\begin{figure}[t]
  \centering
  \includegraphics[width=1.0\linewidth]{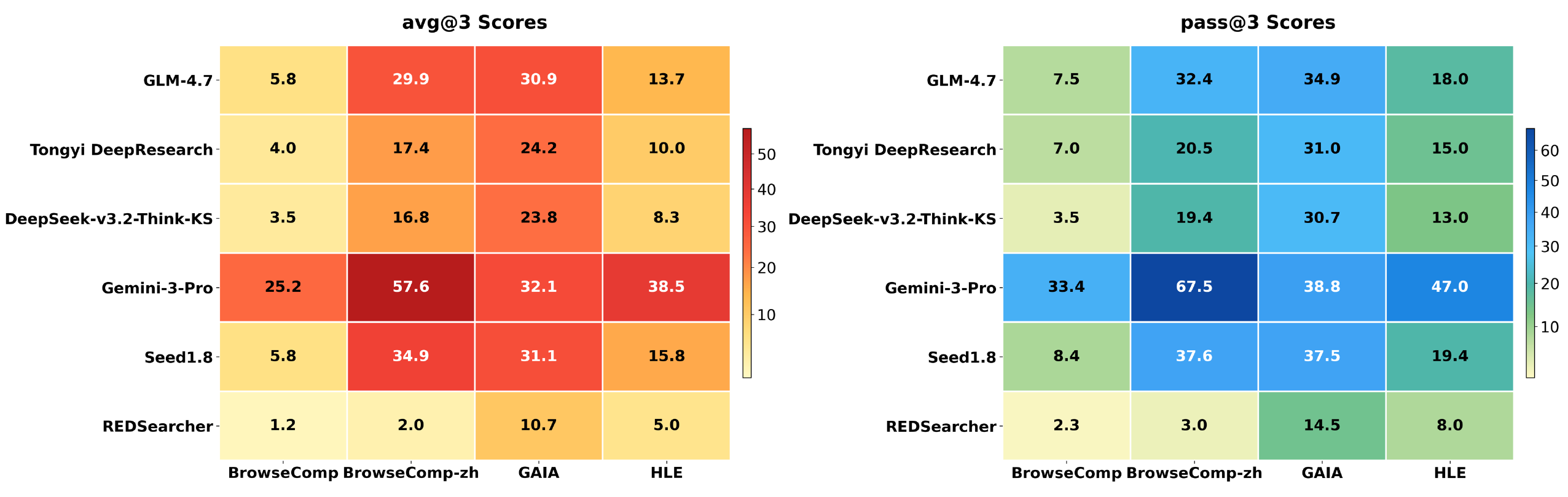}
  \caption{Performance comparison of REDSearcher and existing models in tool-free settings}
  \label{fig:abla_bench}
\end{figure}

Final benchmark accuracy can conflate two factors: success from \emph{tool-mediated evidence acquisition} versus direct recall from \emph{parametric knowledge}. To better isolate tool-use capability, we evaluate each system in two regimes—\emph{tool-free} and \emph{tool-enabled}—and analyze the resulting performance gap (Figure~\ref{fig:abla_bench}). 
In the tool-free regime, \ours{} scores lowest among the compared systems, consistent with reduced reliance on memorized facts or benchmark overlap. When tools are enabled, \ours{} improves substantially and achieves strong overall results, indicating effective planning, evidence gathering, and multi-step synthesis.
Several strong baselines, however, maintain non-trivial accuracy without tools. This may reflect broader pre-training coverage and/or latent benchmark overlap, and can overstate long-horizon, tool-mediated ability if one considers final accuracy alone. Overall, tool-enabled gains provide a more diagnostic signal of deep-search competence by more directly measuring how agents benefit from iterative tool interactions.

\subsection{Multimodal Experimental Results}
\subsubsection{Main Results}
Table~\ref{tab:mm_main} summarizes results on multimodal search benchmarks, where queries and evidence include visual inputs.
Our model delivers strong vision-language search performance, demonstrating effective visual grounding and multimodal evidence integration.
On highly challenging benchmarks such as MM-BrowseComp~\cite{li2025mm}, our method achieves competitive performance against state-of-the-art systems (e.g., Gemini-3-Pro~\cite{gemini3pro2025} and Seed1.8~\cite{seedseed1}), while substantially outperforming a strong Qwen3-VL-235B~\cite{bai2025qwen3vltechnicalreport} agent baseline.
Meanwhile, on relatively simpler multimodal search benchmarks (e.g., MMSearch~\cite{jiang2024mmsearch} and LiveVQA~\cite{fu2025livevqa}), our approach maintains excellent results, indicating robust multimodal retrieval and reasoning across difficulty levels.
Finally, we also evaluate our multimodal search model on text-only benchmarks, where it achieves strong performance, suggesting that the learned search and reasoning capabilities transfer well even without visual inputs. In addition, we find that reinforcement learning further improves model’s overall performance.

\begin{table}[t]
  \centering
  \caption{Main results on multimodal search benchmarks. $\dagger$ denotes results evaluated using the same evaluation tools as ours, and $*$ denotes results taken from~\cite{seedseed1}.}
  \label{tab:mm_main}
  \resizebox{\linewidth}{!}{%
    \renewcommand{\arraystretch}{1.15}
\setlength{\tabcolsep}{4pt}
\begin{tabular}{lcccccc|cccc}
  \toprule
\makecell[l]{Model} & \makecell{Params} & \makecell{MM-Browse\\Comp~\cite{li2025mm}} & \makecell{Browse\\Comp-VL\cite{geng2025webwatcher}} & \makecell{MMSearch\\Plus~\cite{tao2025mmsearch}} & \makecell{MM\\Search~\cite{wu2025mmsearch}} & \makecell{Live\\VQA~\cite{fu2025livevqa}} & \makecell{HLE\\(text)~\cite{phan2025humanity}} & \makecell{HLE\\-VL~\cite{phan2025humanity}} & \makecell{Browse\\Comp~\cite{wei2025browsecomp}} & \makecell{Browse\\Comp-ZH~\cite{zhou2025browsecomp}} \\
  \midrule  
  \rowcolor{gray!20}
\multicolumn{11}{c}{\textbf{\footnotesize Proprietary Deep Research Agents}} \\
  \midrule
  Gemini-2.5-Flash~\cite{comanici2025gemini}           & -- & 5.6 & 44.6   & 19.9     & 64.0    & 73.0  & -- & -- & -- & -- \\
  Gemini-2.5-Pro~\cite{comanici2025gemini}           & -- & 7.1 & 49.9   & 22.2    & 69.0   & 76.0   & - & - & 7.6 & 27.3 \\
  Seed1.8~\cite{seedseed1}                & -- & 46.3 & --   & --   & --   & --   & 40.9 & 31.5 & 67.6 & 81.3 \\
  Seed1.8$^{\dagger}$~\cite{seedseed1}                        & -- & 21.4 & 54.1 & 11.0 & 69.7 & 62.4 & --   & --   & --   & --   \\
  GPT-5~\cite{singh2025openai} & -- & -- & 46.1   & 17.2    & 63.7    & 73.3  & 41.7 & -- & 54.9 & 63.0 \\
  Gemini-3-Pro$^{\dagger}$~\cite{gemini3pro2025}               & -- & 28.5 & 56.4 & 38.1 & 73.0   & 79.9 & 45.8$^{*}$  & 36.0$^{*}$   & 37.8$^{*}$ & 51.6$^{*}$ \\
  \midrule  
  \rowcolor{gray!20}
  \multicolumn{11}{c}{\textbf{\footnotesize Multimodal Agent Flow}} \\
  \midrule
  Qwen2.5-VL~\cite{bai2025qwen2}   & 72B & 1.8 & 10.2 & -   & 29.2 & 35.7 & --   & 4.9   & --  & --  \\
  Qwen3-VL Thinking~\cite{bai2025qwen3vltechnicalreport}  & 30B & 10.7 & 37.1 & 11.0   & 59.7 & 64.8 & 8.8   & 8.7   & 0.2	  & 7.2  \\
  Qwen3-VL Thinking~\cite{bai2025qwen3vltechnicalreport} & 235B & 12.1 & 43.1 & 17.4 & 63.3 & 70.2 & 14.5 & 14.1 & 0.3 & 18.6 \\
  \midrule  
  \rowcolor{gray!20}
  \multicolumn{11}{c}{\textbf{\footnotesize Multimodal Deep Research Agent}} \\
  \midrule
  MMSearch-R1~\cite{wu2025mmsearch}             & 7B & -- & -- & -- & 53.8   & 48.4 & --  & --   & -- & -- \\
  WebWatcher~\cite{geng2025webwatcher}              & 32B & -- & 27.0 & -- & 55.3   & 58.7 & --  & 13.6   & -- & -- \\
  DeepEyesV2~\cite{hong2025deepeyesv2}        & 7B & -- & -- & -- & 63.7   & -- & --  & --   & -- & -- \\
  Vision-DeepResearch~\cite{huang2026vision} & 30B & -- & 53.7 & 28.5 & 69.6  & 77.6 & --  & --   & -- & -- \\
  \midrule
  \cellcolor{pink!30}{REDSearcher-MM-SFT}                          & \cellcolor{pink!30}{30B} & \cellcolor{pink!30}{25.3} &  \cellcolor{pink!30}{55.3} & \cellcolor{pink!30}{20.2} & \cellcolor{pink!30}{70.3}   & \cellcolor{pink!30}{78.5}   & \cellcolor{pink!30}{24.4} & \cellcolor{pink!30}{24.2} & \cellcolor{pink!30}{30.1} & \cellcolor{pink!30}{43.1} \\
  \cellcolor{pink!30}{REDSearcher-MM-RL}                          & \cellcolor{pink!30}{30B} &  \cellcolor{pink!30}{23.5} & \cellcolor{pink!30}{57.2} & \cellcolor{pink!30}{26.6} & \cellcolor{pink!30}{72.9} & \cellcolor{pink!30}{79.3} & \cellcolor{pink!30}{25.3} & \cellcolor{pink!30}{25.6} & \cellcolor{pink!30}{31.2} & \cellcolor{pink!30}{44.5} \\
  \bottomrule
\end{tabular}
  }
\end{table}

\subsubsection{MultiModal DeepReSearch Analysis}

\begin{figure}[ht]
  \centering
  \includegraphics[width=\linewidth]{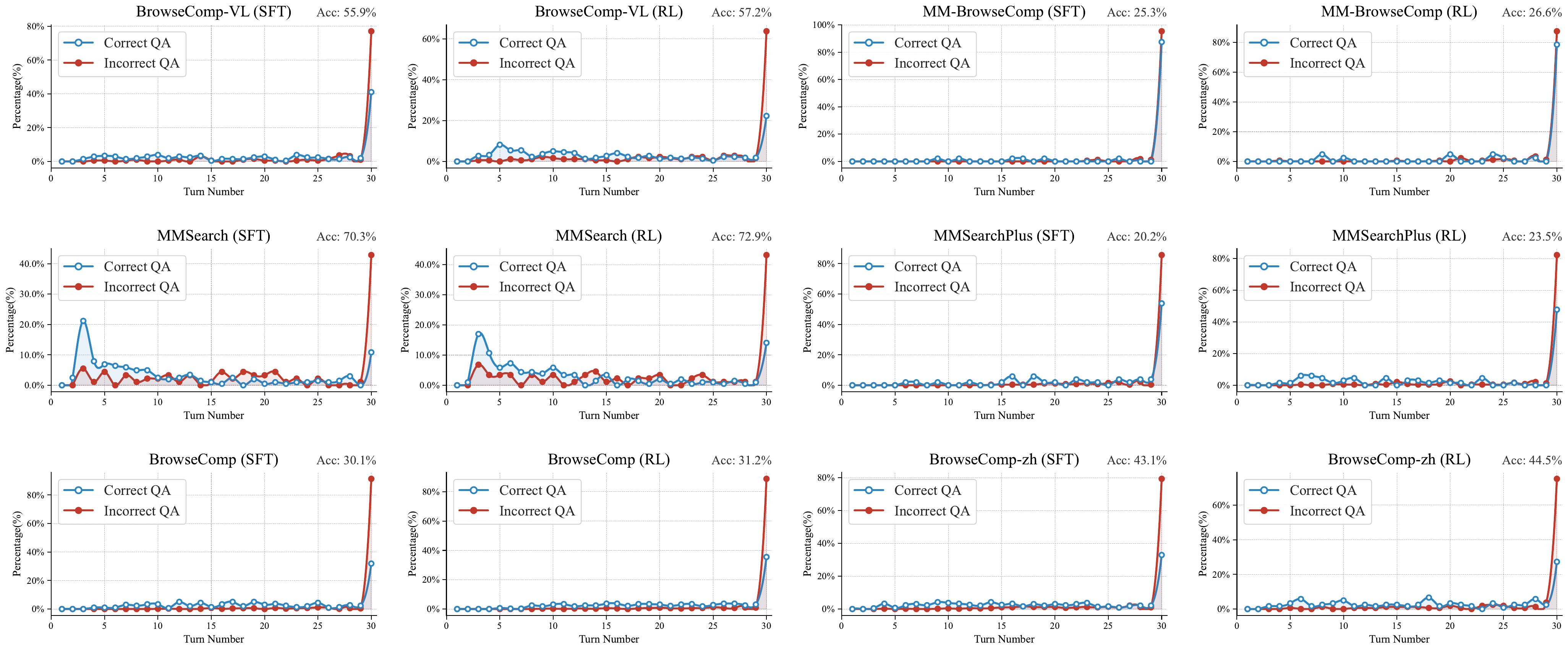}
  \caption{Turns distribution of REDSearcher-MM on different kinds of benchmark.}
  \label{fig:mm_turns}
\end{figure}

\paragraph{Turns Distribution across Different Difficulty Benchmarks.}
We categorize the benchmarks into two groups according to their accuracy and difficulty: \emph{simple} and \emph{challenging}.
We then analyze the distribution of tool-usage turns (i.e., the number of invoked tool calls) for both correct and incorrect predictions in Figure~\ref{fig:mm_turns}.
Note that we enforce a hard cutoff at 30 turns, where the model is forced to output a final answer.
We observe three phenomena from the turn distributions:
(1) The turn distributions differ substantially between the simple and challenging benchmarks: simple benchmarks typically require only a small number of turns for the model to retrieve sufficient evidence and answer with high confidence, whereas challenging benchmarks often demand many more search turns.
(2) The model sometimes continues searching even after it has already encountered the correct evidence, due to insufficient confidence to finalize an answer.
(3) This ``over-searching'' behavior is more pronounced on challenging benchmarks, where a large fraction of examples concentrate near the 30-turn cutoff, indicating that the model frequently keeps searching until it is forced to answer.

Furthermore, we observe a reduction in the number of tool-use turns after RL training, a trend that is particularly pronounced on relatively simple benchmarks. We attribute this to the strict search turn limit (20 turns) imposes during the RL phase, which encourages the model to minimize search steps while maintaining response accuracy.

\begin{figure}[ht]
  \centering
  \includegraphics[width=\linewidth]{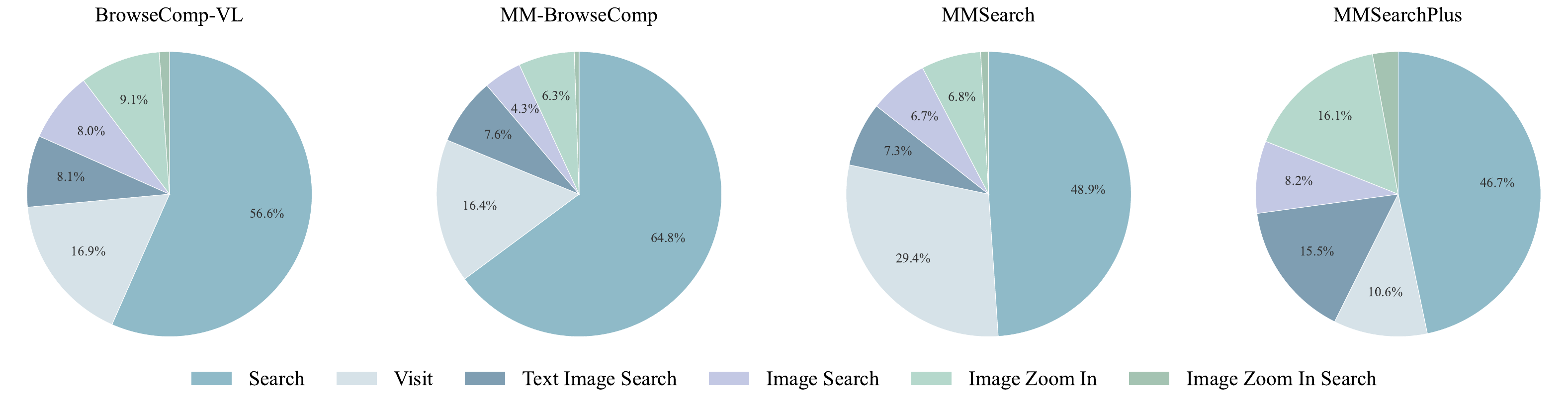}
  \caption{Tool category distribution of REDSearcher-MM.}
  \label{fig:mm_tool_category}
\end{figure}

\paragraph{Tool Category Distribution.}
We further analyze tool usage by categorizing tool calls into different types in Figure~\ref{fig:mm_tool_category}, and we observe clear differences across benchmarks with different characteristics and difficulty.
For example, MMSearch~\cite{jiang2024mmsearch} mainly concentrates on web search and webpage browsing, whereas the more challenging MM-BrowseComp~\cite{li2025mm} induces substantially more text-search steps due to its long-horizon evidence gathering requirements.
In contrast, MMSearch-Plus~\cite{tao2025mmsearch} emphasizes fine-grained visual perception in query construction, which leads to more frequent image-centric operations such as zoom-in and image search.

\begin{figure}[ht]
  \centering
  \includegraphics[width=\linewidth]{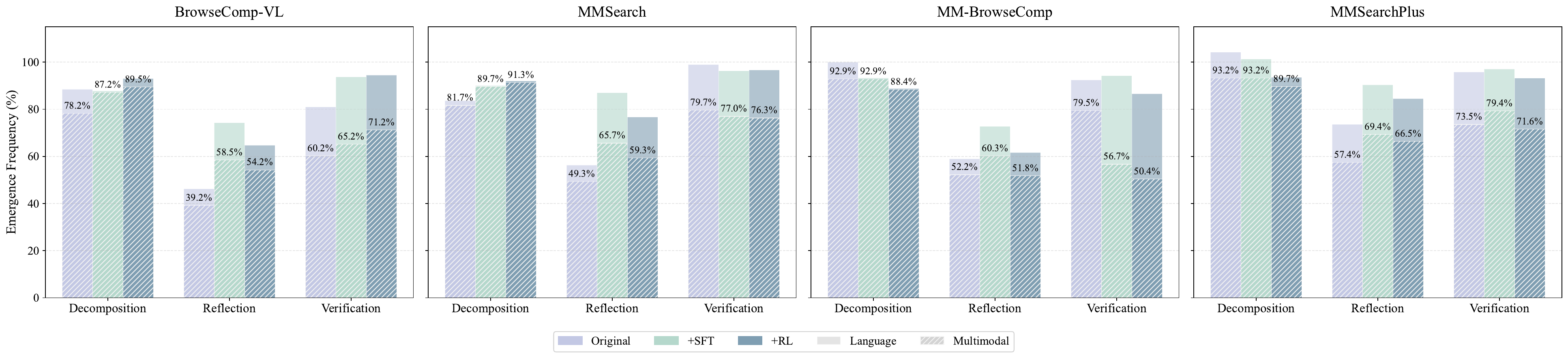}
  \caption{Thinking patterns of REDSearcher-MM on several multimodal search benchmarks.}
  \label{fig:mm_cognitive}
\end{figure}

\paragraph{Thinking Patterns.}
We further characterize the model's high-level thinking patterns during tool use in Figure~\ref{fig:mm_cognitive}, which can be broadly grouped into three types:
(1) \textbf{Decomposition}, where the model breaks a complex query into smaller, actionable sub-questions and solves them sequentially via targeted tool calls;
(2) \textbf{Reflection}, where the model revisits intermediate conclusions, identifies missing evidence or uncertainty, and adjusts the search plan accordingly; and
(3) \textbf{Verification}, where the model cross-checks candidate answers against additional sources (or multiple pieces of evidence) before committing to a final response. It can be seen that model’s thinking patterns differ across benchmarks of varying difficulty levels and types. For relatively simple benchmarks (i.e., BroseComp-VL~\cite{geng2025webwatcher} and MMSearch~\cite{jiang2024mmsearch}), there is less decomposition, a much smaller proportion of reflection, and also a much lower proportion of verification. In addition, on multimodal search benchmarks, the model is more likely to take visual information into account during its reasoning.

\section{Conclusion}

We present \ours{}, a scalable framework for training long-horizon deep search agents across text and multimodal settings. 
To address the scarcity of high-quality training data, we propose dual-constrained task synthesis that generates structurally complex reasoning tasks with dispersed evidence, ensuring the necessity of iterative planning and cross-document synthesis. 
To reduce the computational and temporal costs of trajectory collection, we introduce cost-efficient mid-training that separates atomic subskill acquisition from interactive execution, combined with a functionally equivalent simulation environment that enables high-throughput trajectory generation without relying on expensive live API calls. 
Building upon this foundation, we advance the model's search intelligence through trajectory synthesis, supervised fine-tuning, and agentic reinforcement learning. 
Together, these contributions provide a practical pathway for scaling deep search agents, marking a significant step toward transforming LLMs from passive knowledge retrievers into proactive agents capable of long-horizon reasoning and autonomous exploration over the open world.

\clearpage

\clearpage
\section*{Contributions}

\vspace{0.5cm}

\noindent \textbf{Core Contributors} \\
Zheng Chu${}^1$, Xiao Wang${}^2$, Jack Hong${}^2$

\vspace{0.5cm}

\noindent \textbf{Contributors} \\
Huiming Fan${}^1$, Yuqi Huang${}^3$, Yue Yang${}^3$, Guohai Xu${}^2$, Chenxiao Zhao${}^2$, Cheng Xiang${}^2$, \\ Shengchao Hu${}^3$, Dongdong Kuang${}^2$, Bing Qin${}^1$, Xing Yu${}^2$

\vspace{0.5cm}

\noindent \textbf{Project Leader} \\
Xiao Wang${}^{2}$

\vspace{0.5cm}

\noindent \textbf{Advisors} \\
Ming Liu${}^1$, Xiao Wang${}^2$

\vspace{1.5cm} 

\noindent \rule{3cm}{0.4pt} \\ 
\vspace{0.2cm}

\noindent {\footnotesize
${}^1$ Harbin Institute of Technology \\
${}^2$ Xiaohongshu Inc. \\
${}^3$ Shanghai JiaoTong University
}

{\footnotesize \texttt{Emails:} \texttt{zchu@ir.hit.edu.cn}, \texttt{wangxiao14@xiaohongshu.com}}, \texttt{mliu@ir.hit.edu.cn} 

\clearpage









{\small
\bibliographystyle{plain}
\bibliography{references}
}

\newpage
\appendix

\section{Implementation Details.}
\label{Sec:Implementation details}
\ours{} are trained based on Qwen3-30B-A3B~\cite{yang2025qwen3}.
During the mid-training phase, we use a batch size of 512 in Stage 1 and a batch size of 256 in Stage 2.
For the SFT stage, we use a batch size of 128.
Throughout the mid-training and SFT phases, the learning rate decays from 5e-5 to 1e-6, with a linear decay in mid-training followed by cosine decay in sft.
We adopt GRPO as RL training algorithm. 
Each mini-step consists of 32 queries, with 16 rollout samples per query, resulting in a mini-batch size of 512. 
The learning rate is fixed at 1e-6 throughout this stage. 
We set clip high to 0.28, and do not use entropy loss and kl loss.
We employ Truncated Important Sampling (TIS) and Routing Replay (R2) to mitigate inconsistency issues.
To ensure stable gradient updates during RL training, we filter out abnormal samples that exhibit repetition, excessive length, or frequent tool call failures. These samples still participate in advantage computation but are excluded from gradient updates.
During inference, we set the temperature to 0.85, top\_p to 0.95, and the maximum length to 128K.
Once the model exceeds the context limit, we roll back to the previous round and force an answer.
For summarizer used in visit tool, we employ Qwen3-30B-A3B-Instruct-2507.
For LLM-as-Judge, we use GPT-OSS-120B.

For multimodal search, we use Qwen3-VL-30B-A3B-Thinking~\cite{bai2025qwen3vltechnicalreport}.
For SFT, we train with a batch size of 128 and a learning rate of $1\times 10^{-5}$.
The model is optimized for three epochs using the AdamW optimizer with cosine learning-rate decay.
For RL, we adopt GRPO~\cite{guo2025deepseek,shao2024deepseekmath} as the optimization algorithm, with a batch size of 32 and 8 rollouts per prompt.
The KL coefficient is set to 0.0, and the maximum response length is capped at 32{,}768 tokens.
During RL, we cap the tool-calling horizon to a maximum of 20 tool calls per episode.

\clearpage
\section{System Prompt}

\begin{tcolorbox}[title=System Prompt of REDSearcher]
You are a deep search assistant. Your primary role is to perform rigorous, multi-step, multi-source investigations on any topic—covering both broad, open-domain questions and highly specialized academic inquiries. 
\\

For each user request, you must actively seek out and cross-check information from credible and diverse sources, then integrate the findings into a response that is comprehensive, accurate, well-structured, and objective.
\\

\#\# Operating principle
\\
1. **Plan and execute research**: Break complex questions into sub-questions, gather evidence across multiple sources, and prioritize primary sources and authoritative references when available.
\\
2. **Evaluate source quality**: Prefer reputable institutions, peer-reviewed research, official documentation, and high-quality journalism. Note uncertainty, conflicts, and limitations when sources disagree.
\\
3. **Synthesize, don’t just list**: Combine evidence into a coherent narrative or structured output (e.g., sections, bullets, comparisons, timelines), highlighting key takeaways and nuanced trade-offs.
\\
4. **Maintain neutrality**: Present competing viewpoints fairly when relevant, and avoid unsupported speculation.
\\

When you have collected sufficient information and are ready to deliver the definitive response, you must wrap the entire final answer in **\textcolor{black}{\textbf{<answer></answer>}}** tags.
\\

\# Tools
\\

You may call one or more functions to assist with the user query.
\\

You are provided with function signatures within \textcolor{black}{\textbf{<tools></tools>}} XML tags:\\
\textcolor{black}{\textbf{<tools>}}\\

\{"type": "function", "function": \{"name": "\textcolor{red1}{\textbf{search}}", "description": "Perform Google web searches then returns a string of the top search results. Accepts multiple queries.", "parameters": \{"type": "object", "properties": \{"query": \{"type": "array", "items": \{"type": "string", "description": "The search query."\}, "minItems": 1, "description": "The list of search queries."\}\}, "required": ["query"]\}\}\}
\\
\{"type": "function", "function": \{"name": "\textcolor{red1}{\textbf{visit}}", "description": "Visit webpage(s) and return the summary of the content.", "parameters": \{"type": "object", "properties": \{"url": \{"type": "array", "items": \{"type": "string"\}, "description": "The URL(s) of the webpage(s) to visit. Can be a single URL or an array of URLs."\}, "goal": \{"type": "string", "description": "The specific information goal for visiting webpage(s)."\}\}, "required": ["url", "goal"]\}\}\}

\{"type": "function", "function": \{"name": "\textcolor{red1}{\textbf{PythonInterpreter}}", "description": "Executes Python code in a sandboxed environment. To use this tool, you must follow this format:1. The 'arguments' JSON object must be empty: \{\}.2. The Python code to be executed must be placed immediately after the JSON block, enclosed within <code> and </code> tags.IMPORTANT: Any output you want to see MUST be printed to standard output using the print() function.Example of a correct call:<\texttt{tool\_call}>\{"name": "PythonInterpreter", "arguments": \{\}\}<code>import numpy as np \# Your code here print(f"The result is: {np.mean([1,2,3])}") </code></\texttt{tool\_call}>", "parameters": \{"type": "object", "properties": \{\}, "required": []\}\}\}

\{"type": "function", "function": \{"name": "\textcolor{red1}{\textbf{\texttt{google\_scholar}}}", "description": "Leverage Google Scholar to retrieve relevant information from academic publications. Accepts multiple queries.", "parameters": \{"type": "object", "properties": \{"query": \{"type": "array", "items": \{"type": "string", "description": "The search query."\}, "minItems": 1, "description": "The list of search queries for Google Scholar."\}\}, "required": ["query"]\}\}\}

\end{tcolorbox}

\begin{tcolorbox}[title=System Prompt of REDSearcher, breakable, enhanced]

\{"type": "function", "function": \{"name": "\textcolor{red1}{\textbf{\texttt{google\_maps}}}", "description": "Search Google Maps places. Returns a list of places with name, address, coordinates, ratings, categories, opening hours, and place identifiers.", "parameters": \{"type": "object", "properties": \{"q": \{"type": "string", "description": "Google Maps search query."\}, "page": \{"type": "integer", "description": "Page number of results.", "default": 1, "minimum": 1\}\}, "required": ["q"]\}\}\}\\

For each function call, return a json object with function name and arguments within \textcolor{black}{\textbf{<tool\_call></tool\_call>}} XML tags:
\\
\textcolor{black}{\textbf{<tool\_call>}}\\
\{"name": <function-name>, "arguments": <args-json-object>\}\\
\textcolor{black}{\textbf{</tool\_call>}}
\\
\end{tcolorbox}

\begin{tcolorbox}[title=System Prompt of REDSearcher-MM]
You are an agent - please keep going until the user's query is completely resolved, before ending your turn and yielding back to the user. Only terminate your turn when you are sure that the problem is solved.
\\

Solve the following problem step by step. If you find you don't have sufficient knowledge to confidently answer the question, you MUST conduct search to thoroughly seek the internet for information. No matter how complex the query, you will not give up until you find the corresponding information.
\\

You can conduct image search, which will trigger a Google Lens search using the original image to retrieve relevant information that can help you confirm the visual content, and text search, which will use Google Search to return relevant information based on your query.
\\

You MUST plan extensively before each function call, and reflect extensively on the outcomes of the previous function calls. DO NOT do this entire process by making function calls only, as this can impair your ability to solve the problem and think insightfully.
\\

For all the provided images, in order, the i-th image has already been read into the global variable \texttt{`image\_i'}
using the \texttt{`PIL.Image.open()'} function. For example, the first image can be accessed as \texttt{`image\_0'}. When writing Python code, you can directly use these variables. without needing to read them again.
\\

All image-capable tools also accept an optional JSON argument \texttt{`\{"image\_index": k, ...\}'} to explicitly choose which image to operate on. The index is zero-based, so \texttt{`0'} refers to the first image by default; if you omit this field, the tool automatically uses \texttt{`image\_0'}.
\\

Please put the answer within \textcolor{black}{\textbf{<answer></answer>}} tags. Inside these tags, the core answer must be wrapped in \texttt{\textbackslash boxed\{\}}. In addition, include supporting evidence, explanations, and broader context within the same \textcolor{black}{\textbf{<answer></answer>}} section.
\\

\# Tools
\\

You may call one or more functions to assist with the user query.
\\

You are provided with function signatures within \textcolor{black}{\textbf{<tools></tools>}} XML tags:\\
\textcolor{black}{\textbf{<tools>}}\\
\{"type": "function", "function": \{"name": "\textcolor{red1}{\textbf{\texttt{text\_search}}}", "description": "A web search tool that returns a list of web pages with summaries based on input queries. Queries should be concise and clear. Break down complex questions into multiple steps. If no useful results are found, adjust the query by reducing qualifiers or changing the search approach. For better results, use Chinese queries for Chinese resources and English for non-Chinese resources.", "parameters": \{"type": "object", "properties": \{"query": \{"type": "array", "items": \{"type": "string", "description": "The search query."\}, "minItems": 1, "description": "List of search queries."\}\}, "required": ["query"]\}\}\}\\
\{"type": "function", "function": \{"name": "\textcolor{red1}{\textbf{\texttt{text\_image\_search}}}", "description": "A web image search tool that returns information such as images and corresponding original webpage URLs based on input queries. Queries should be concise and clear. Break down complex questions into multiple steps. If no useful results are found, adjust the query by reducing qualifiers or changing the search approach. For better results, use Chinese queries for Chinese resources and English for non-Chinese resources.", "parameters": \{"type": "object", "properties": \{"query": \{"type": "string", "description": "search queries"\}\}, "required": ["query"]\}\}\}

\end{tcolorbox}
\begin{tcolorbox}[title=System Prompt of REDSearcher-MM, breakable, enhanced]

\{"type": "function", "function": \{"name": "\textcolor{red1}{\textbf{\texttt{image\_zoom\_in\_search}}}", "description": "This is a visual search tool used to search the entire web for image results similar to the input image. It performs searches based on images and returns relevant images, videos, and product information.", "parameters": \{"type": "object", "properties": \{"\texttt{image\_index}": \{"type": "integer", "description": "Optional index of the image to search (default 0)."\}, "\texttt{bbox\_2d}": \{"type": "array", "items": \{"type": "integer"\}, "description": "ROI coordinates. **Must** be an array containing 4 integer values (ranging from 0 to 1000), representing the normalized coordinates of the cropping region. Format. [\texttt{left\_top\_x}, \texttt{left\_top\_y}, \texttt{right\_bottom\_x}, \texttt{right\_bottom\_y}]. Example. [100, 289, 381, 465]. The meaningless full-image coordinates like [0, 0, 1000, 1000] are prohibited."\}\}, "required": ["\texttt{image\_index}", "\texttt{bbox\_2d}"]\}\}\}

\{"type": "function", "function": \{"name": "\textcolor{red1}{\textbf{\texttt{image\_search}}}", "description": "This is a visual search tool used to search the entire web for image results similar to the input image. It performs searches based on images and returns relevant images, videos, and product information.", "parameters": \{"type": "object", "properties": \{"\texttt{image\_index}": \{"type": "integer", "description": "Optional index of the image to search (default 0)."\}\}, "required": ["\texttt{image\_index}"]\}\}\}

\{"type": "function", "function": \{"name": "\textcolor{red1}{\textbf{\texttt{web\_summary}}}", "description": "This is a web summary tool that can open links and summarize all relevant information on the page according to the goal. It is recommended to invoke this tool for valuable links to obtain information. Valuable links include but are not limited to the following types. 1. URLs explicitly provided in the task; 2. URLs with relevant abstracts returned by search results (source URLs of images returned by image searches are also included in this scope); 3. URLs contained in the content returned by previous calls of \texttt{web\_summary} and judged to potentially contain useful information. Please try to avoid constructing links out of thin air.", "parameters": \{"type": "object", "properties": \{"url": \{"type": "string", "description": "The target link shall be a complete URL (starting with http)."\}, "goal": \{"type": "string", "description": "Requirement description text, which elaborates on the content to be retrieved from the current URL in detail."\}\}, "required": ["url", "goal"]\}\}\}

\{"type": "function", "function": \{"name": "\textcolor{red1}{\textbf{\texttt{image\_zoom\_in}}}", "description": "Image zoom-in tool. Crop and magnify a designated area for viewing image details, text recognition or subsequent identification processes. It is extremely useful for object location in images and content retrieval from images. For the thinking section after ZoomIn, a detailed interpretation of the zoomed-in image is required.", "parameters": \{"type": "object", "properties": \{"\texttt{image\_index}": \{"type": "integer", "description": "Optional index of the image to search (default 0)."\}, "\texttt{bbox\_2d}": \{"type": "array", "items": \{"type": "integer"\}, "description": "ROI coordinates. **Must** be an array containing 4 integer values (ranging from 0 to 1000), representing the normalized coordinates of the cropping region. Format. [\texttt{left\_top\_x}, \texttt{left\_top\_y}, \texttt{right\_bottom\_x}, \texttt{right\_bottom\_y}]. Example. [100, 289, 381, 465]. The meaningless full-image coordinates like [0, 0, 1000, 1000] are prohibited."\}, "label": \{"type": "string", "description": "Optional labels for describing the content of the cropped area, helping the model better understand the semantic information of the cropped image."\}\}, "required": ["\texttt{image\_index}", "\texttt{bbox\_2d}", "label"]\}\}\}\\
\textcolor{black}{\textbf{<tools>}}\\

For each function call, return a json object with function name and arguments within \textcolor{black}{\textbf{<tool\_call></tool\_call>}} XML tags:
\\
\textcolor{black}{\textbf{<tool\_call>}}\\
\{"name": <function-name>, "arguments": <args-json-object>\}\\
\textcolor{black}{\textbf{</tool\_call>}}
\\
\end{tcolorbox}

\clearpage

\section{Synthetic Data Case}

We show some cases of our synthesized data.

\centering
\framebox{%
\begin{minipage}{15cm}
\ttfamily\footnotesize
{\color{blue}Question:}\\
An industrial entity in the music production sector—a record pressing plant and label—is located approximately 360 km southwest of the city that hosted the 2016 Summer Olympics, within the South American country that experienced a significant outbreak of a mosquito-borne viral disease, colloquially named for a racial stereotype regarding attraction to people of East Asian descent, beginning in late 2016 and intensifying in mid-2017. Its operational commencement coincided with this disease outbreak in July 2017. The founder, whose given name is a common French masculine name and surname is of Hebrew origin meaning 'gift', previously created a limited edition album of 500 copies released three years before the plant's founding. The plant presses records using a format material derived from a flexible, partially crystalline polymer. An example release using this specific material is a limited edition EP of 500 copies created by a musical artist known for the 'psychedelic garage acid punk' genre, who formed in 2005 and is signed to a record label whose acronym HFTG could also refer to a high school metal band known as 'Hanging from the Gallows'. Based on these clues, what is the name of this pressing plant and label, which combines the Portuguese word for vinyl with the name of a South American country?\\
{\color{purple}Answer:}\\
\textbf{Vinil Brasil}
\end{minipage}
}

\centering
\framebox{%
\begin{minipage}{15cm}
\ttfamily\footnotesize
{\color{blue}Question:}\\
During the year the WHO declared COVID-19 a pandemic, a 100-bed healthcare facility located in a suburb approximately 27 km northeast of the financial capital of the Indian state that borders six other states including Gujarat and Madhya Pradesh, experienced a critical generator failure. The failure, a fire caused by a short circuit, occurred about three minutes after sunset and resulted in a fatal evacuation. This incident shared its calendar year with a major electrical fire at a repurposed hotel COVID facility in a city situated on the banks of the Krishna River, approximately 63 km northwest of a major port on the Bay of Bengal coast in the same state. Based on this interconnected timeline of infrastructure failures, what is the identity of the suburban facility where the generator failure proved fatal?\\
{\color{purple}Answer:}\\
\textbf{Apex Hospital}
\end{minipage}
}

\clearpage

\begin{center}
\framebox{%
\begin{minipage}{15cm}
\ttfamily\footnotesize
{\color{blue}Question:}\\
Held on a one-mile oval track in the United States during the final months of 2001, this professional stock car racing event featured a pole winner born at the start of the 1980s who had previously set a qualifying record while still a teenager. Identify the official name of the competition, which was won by the driver of the vehicle displaying the specific livery shown in the provided image.\\
\includegraphics[width=\linewidth]{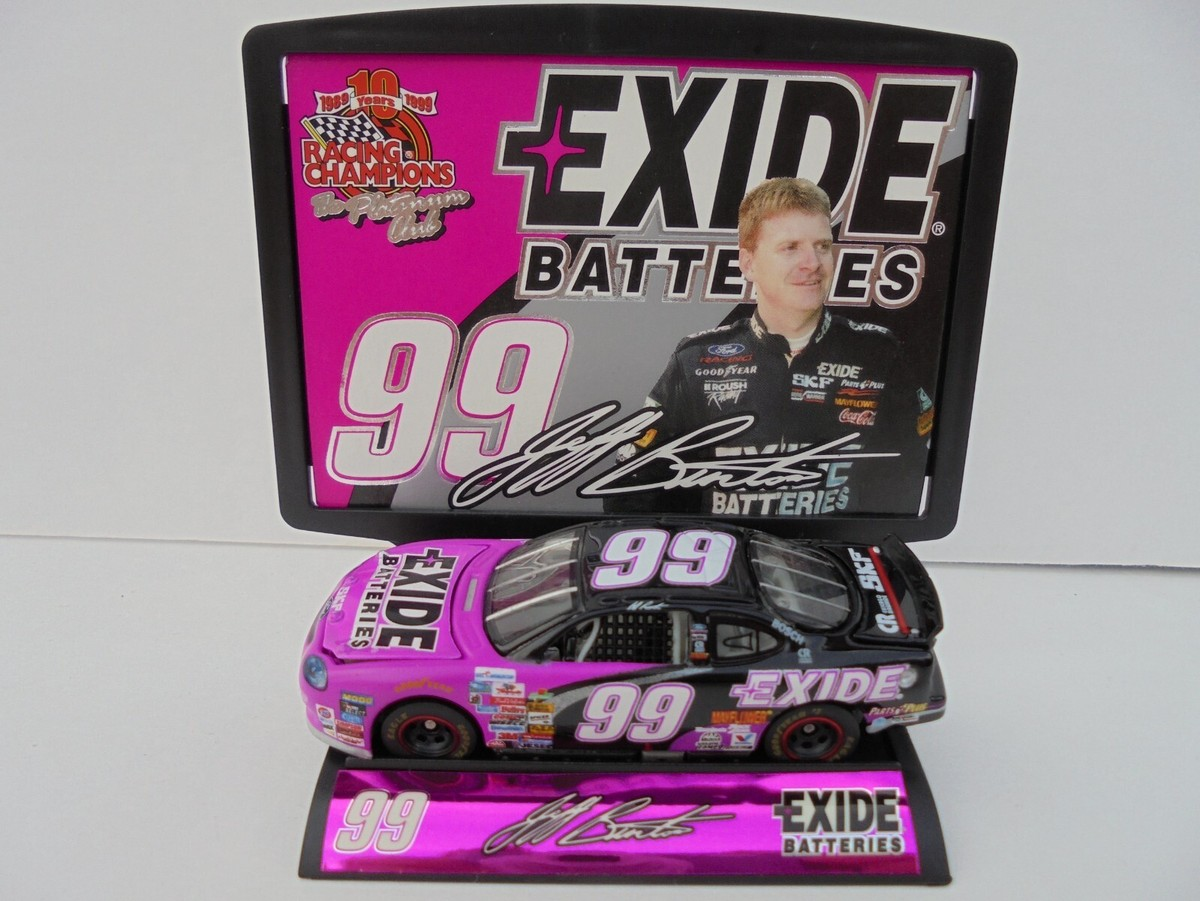}\\
{\color{purple}Answer:}\\
\textbf{14th Annual Checker Auto Parts}
\end{minipage}
}
\end{center}

\begin{center}
\framebox{%
\begin{minipage}{15cm}
\ttfamily\footnotesize
{\color{blue}Question:}\\
In the late 1940s, a strategic hilltop village was depopulated during a military operation. This site is situated approximately halfway between the historic market town containing the medieval tower shown in the image and a globally revered holy metropolis to the east. Its lands are currently occupied by a modern cooperative settlement—located in a time zone two hours ahead of UTC—whose name translates to 'Root' or 'Source', a reference derived from the Septuagint translation of the Book of Joshua. Identify the name of the depopulated village.
\\
\includegraphics[width=\linewidth]{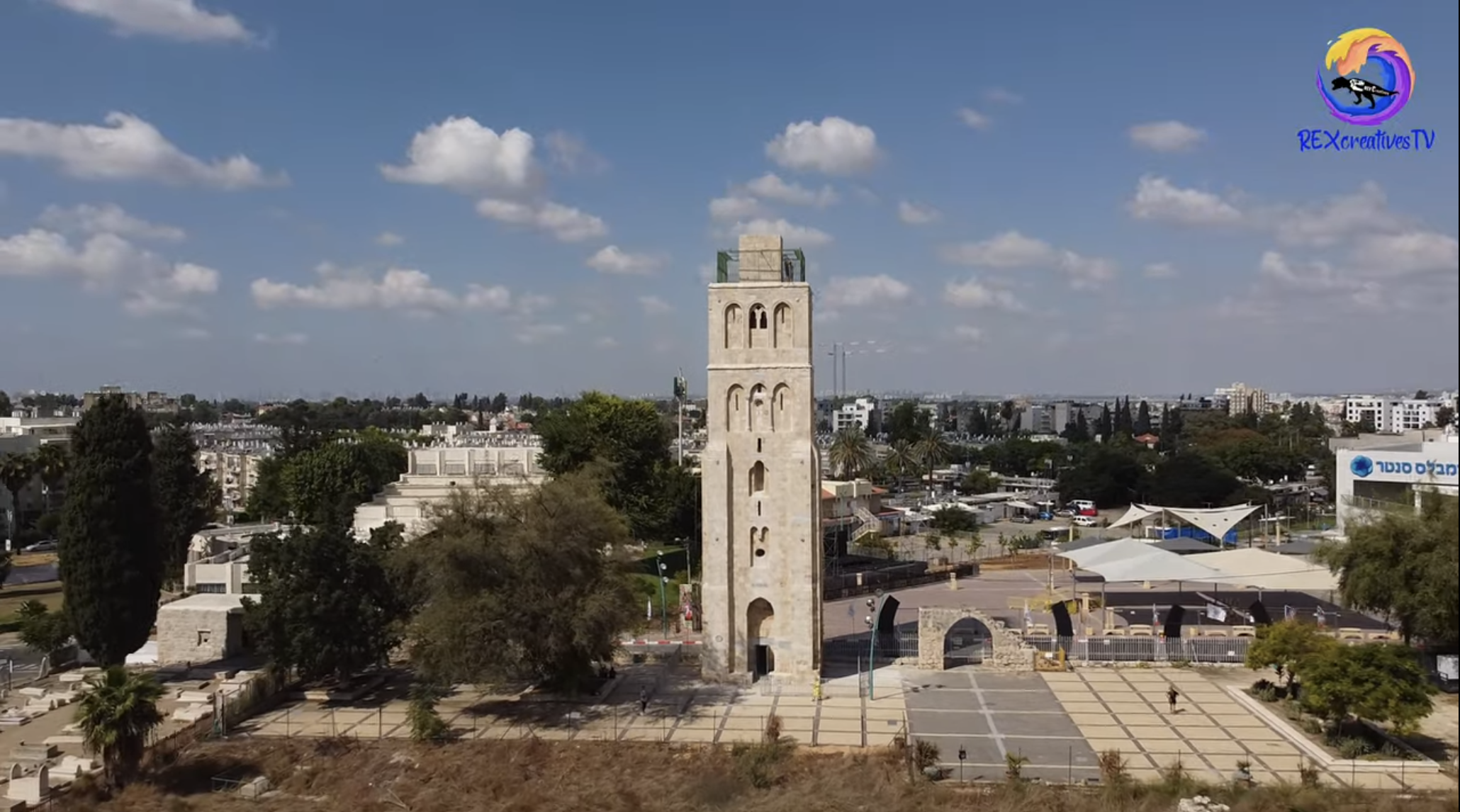}\\
{\color{purple}Answer:}\\
\textbf{Saris}
\end{minipage}
}
\end{center}

\begin{center}
\framebox{%
\begin{minipage}{15cm}
\ttfamily\footnotesize
{\color{blue}Question:}\\
The career of this gridiron athlete began in the metropolitan area defined by the massive copper sculpture shown in the image. He attended a secondary school in a neighboring district—an institution established to relieve overcrowding in the same year a major volcanic eruption occurred less than 100 miles away, and which shares its name with a historic local settlement. Identify the athlete.
\\
\includegraphics[width=\linewidth]{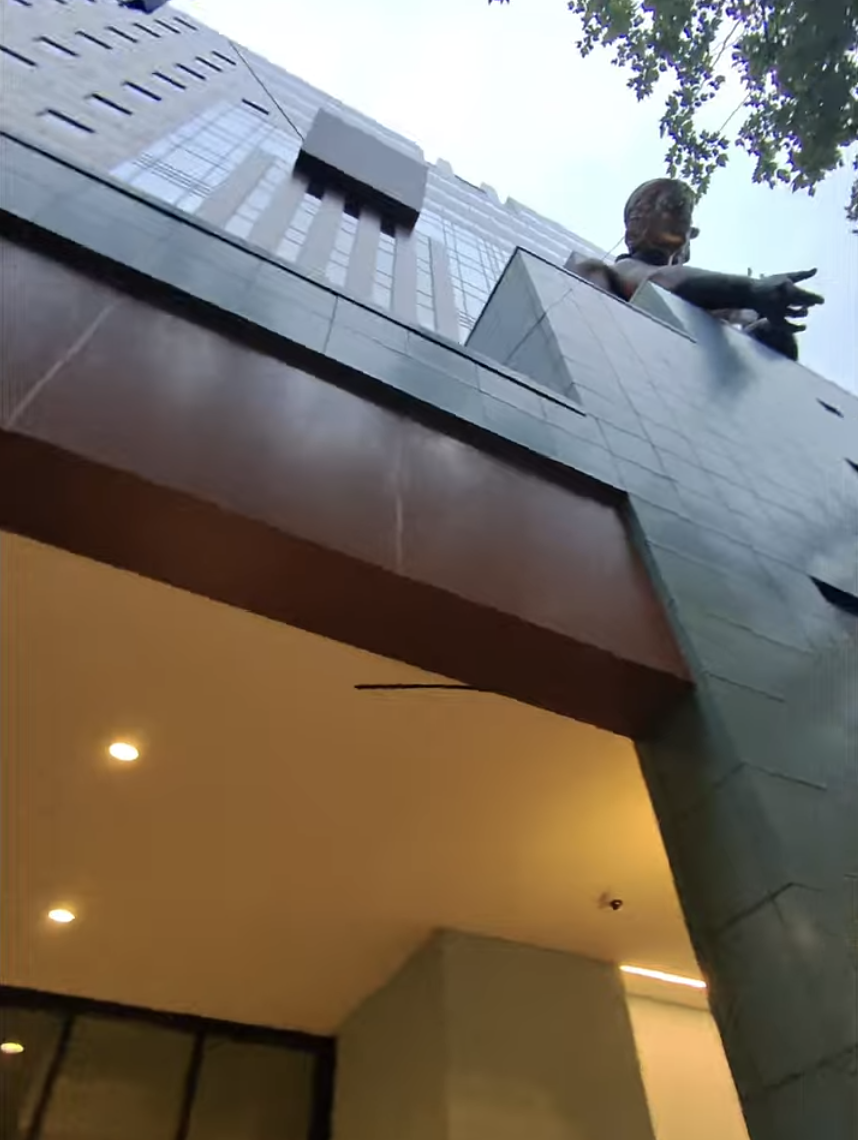}\\
{\color{purple}Answer:}\\
\textbf{Erik Ainge}
\end{minipage}
}
\end{center}

\end{document}